\documentclass{article} % For LaTeX2e
\usepackage{hyperref}       % hyperlinks
\usepackage{url}            % simple URL typesetting
\usepackage{booktabs}       % professional-quality tables
\usepackage{amsfonts}       % blackboard math symbols
\usepackage{amsmath}
\usepackage[numbers]{natbib2}

\usepackage{amsthm}
\usepackage{amssymb}
\usepackage{nicefrac}       % compact symbols for 1/2, etc.
\usepackage{microtype}      % microtypography
\usepackage{graphicx}
\usepackage{tabularx, booktabs, makecell, caption}
\usepackage{siunitx}
\usepackage{multirow}
\usepackage{enumitem}
\usepackage{color}
\usepackage{wrapfig}
\usepackage{algorithm}
\usepackage{algorithmic}
\usepackage{mwe}
\usepackage{graphbox}
\usepackage{wrapfig}
\usepackage{authblk}
\usepackage{fullpage}

\DeclareMathOperator*{\argmin}{arg\,min}

\newtheorem{thm}{Theorem}

\newtheorem{lemma}{Lemma}
\newtheorem{cor}{Corollary}

\newtheorem{definition}{Definition}
\newtheorem{remark}{Remark}

\newtheorem{assume}{Assumption}

\newcommand{\myparagraph}[1]{\textbf{#1}}

\title{Training Federated GANs with Theoretical Guarantees: A Universal Aggregation Approach}

\author[1]{Yikai Zhang \thanks{equal contribution}}
\author[1]{Hui Qu$^*$ }
\author[1]{Qi Chang$^*$ }
\author[2]{Huidong Liu$^*$ }
\author[1]{Dimitris Metaxas}
\author[3]{Chao Chen}
\affil[1]{Department of Computer Science, Rutgers University}
\affil[ ]{ \{yz422, hq43,qc58,dnm\}@cs.rutgers.edu}
\affil[2]{Deparment of Computer Science, Stony Brook University}
\affil[ ]{ huidliu@cs.stonybrook.edu}
\affil[3]{Deparment of Biomedical Informatics, Stony Brook University}
\affil[ ]{chao.chen.cchen@gmail.com}

\begin{document}
\maketitle
% \vspace{-1.5em}
\begin{abstract}
Recently, Generative Adversarial Networks (GANs) have demonstrated their potential in federated learning, i.e., learning a centralized model from data privately hosted by multiple sites. A federated GAN jointly trains a centralized generator and multiple private discriminators hosted at different sites. A major theoretical challenge for the federated GAN is the heterogeneity of the local data distributions. Traditional approaches cannot guarantee to learn the target distribution, which is a mixture of the highly different local distributions. This paper tackles this theoretical challenge, and for the first time, provides a provably correct framework for federated GAN. We propose a new approach called Universal Aggregation, which simulates a centralized discriminator via carefully aggregating the mixture of all private discriminators. We prove that a generator trained with this simulated centralized discriminator can learn the desired target distribution. Through synthetic and real datasets, we show that our method can learn the mixture of largely different distributions where existing federated GAN methods fail.  ~\footnote{code available at: https://github.com/yz422/UAGAN}
\end{abstract}

% \vspace{-.7em}
\section{Introduction}
Generative Adversarial Networks (GANs) have attracted much attention due to their ability to generate realistic-looking synthetic data \cite{goodfellow2014generative, zhang2018metagan, liu2019few, shaham2019singan, dai2017good, kumar2017semi}. % \cc{Add GAN applications.} 
In order to obtain a powerful GAN model, one needs to use data with a wide range of characteristics~\cite{qi2019loss}.
% \cc{The first two sentences are not very convincing. Need to polish: why GAN needs more diversified data?}
% to improve the performance of trained model \cite{domingos2012few, dupin2011effects,tsangaratos2016comparison}.
However, these diverse data are often owned by different sources, and to acquire their data is often infeasible. 
For instance, most hospitals and research institutions are unable to share data with the research community, due to privacy concerns \cite{annas2003hipaa,mercuri2004hipaa,milieu2014overview,gostin2009beyond} 
% \cite{annas2003hipaa,centers2003hipaa,mercuri2004hipaa,gostin2009beyond} 
and government regulations~\cite{kerikmae2017challenges,seddon2013cloud}.
% many hospitals and research institutions are wary of cloud platforms and prefer to use their own servers. 

To circumvent the barrier of data sharing for GAN training, one may resort to Federated Learning (FL), a promising new decentralized learning paradigm \cite{mcmahan2017communication}.
In FL, one trains a centralized model but only exchanges model information with different data sources. Since the central model has no direct access to data at each source, privacy concerns are alleviated \cite{yang2019federated,kairouz2019advances}.
This opens the opportunity for a \emph{federated GAN}, i.e., a centralized generator with  multiple local and privately hosted discriminators~\cite{hardy2019md}. 
Each local discriminator is only trained on its local data and provides feedback to the generator w.r.t.~synthesized data (e.g., gradient).
A federated GAN empowers GAN with much more diversified data without violating privacy constraints.
% A local discriminator serves as a shield protecting sensitive data from a querier; it is contained in an independent data center and only has access to local data. \cc{I think this sentence is redundant. I tend to remove it.}
	
Despite the promises, a convincing approach for training a federated GAN remains unknown. 
The major challenge comes from the non-identical local distributions from multiple data sources/entities. The centralized generator is supposed to learn a mixture of these local distributions from different entities, whereas each discriminator is only trained on local data and learns one of the local distributions. The algorithm and theoretical guarantee of traditional single-discriminator GAN  \cite{goodfellow2014generative} do not easily generalize to this federated setting. 
A federated GAN should integrate feedback from local discriminators in an intelligent way, so that the generator can `correctly' learn the mixture distribution.
%  A natural question with such constraint is:
%     training a federated GAN requires aggregating feedbacks from multiple discriminators. 
%         % \vspace{-.1em}
%         \begin{center}
%         \textit{How do we correctly utilize the feedback from multiple privately hosted discriminators? }
%         % \cc{Domestic? Local? Private?}}
%         \end{center}
%     % \vspace{-.1em}
    % % \vspace{-.1em}
    %     $$\vspace{-.1em}\textit{How do we correctly utilize the feedback from multiple domestic discriminators?}\vspace{-.1em}$$
    % % \vspace{-.1em}
    % \cc{Naive sounds bad, use a less offensive word, e.g., natural?} 
    Directly averaging feedbacks from local discriminators~\cite{hardy2019md} results in a strong bias toward common patternsowever, such non-identical distribution setting is classical in federated learning ~\cite{zhao2018federated,smith2017federated,qu2020learn} and characteristic of local data improves the diversity of data. 
    % ; the generator cannot learn distinct features specific to different entities (see Fig.~\ref{fig:arch1}(left)). 
    % % H

%  \vspace{-1em}
    
 In this paper, we propose \emph{the first theoretically guaranteed federated GAN}, that can correctly learn the mixture of local distributions. Our method, called Universal Aggregation GAN (UA-GAN), focuses on the odds value rather than the predictions of local discriminators. We simulate an unbiased centralized discriminator whose odds value approximates that of the mixture of local discriminators. We prove that by aggregating gradients from local discriminators based on the odds value of the central discriminator, we are guaranteed to learn the desired mixture of local distributions.

        % Based on the Jensen-Shannon Entropy divergence \cite{goodfellow2014generative, lin1991divergence}, the UA framework is able to collect feedbacks from multiple discriminators with \emph{provable guarantees}. 
        
A second theoretical contribution of this paper is an analysis of the quality of the federated GAN when the local discriminators cannot perfectly learn with local datasets. This is a real concern in a federated learning setting;  the quantity and quality of local data can be highly variant considering the limitation of real-world institutions/sites.
Classical theoretical analysis of GAN \cite{goodfellow2014generative} assumes an optimal discriminator. To understand the consequence of suboptimal discriminators, 
    we develop a novel analysis framework of the Jensen-Shannon Divergence loss \cite{goodfellow2014generative,lin1991divergence} through the odds value of the local discriminators. We show that when the local discriminators behave suboptimally, the approximation error of the learned generator deteriorates linearly to the error. 

    It is worth noting that our theoretical result on suboptimality also applies to the classical GAN. To the best of our knowledge, \emph{this is the first suboptimality bound on the federated or classical GAN}.
 \begin{figure}
% \begin{figure}[]
% \vspace{-.6em}
% \begin{minipage}[t]{1\linewidth}
    \centering
    % \vspace{-.6em}
    \includegraphics[width=.5\linewidth]{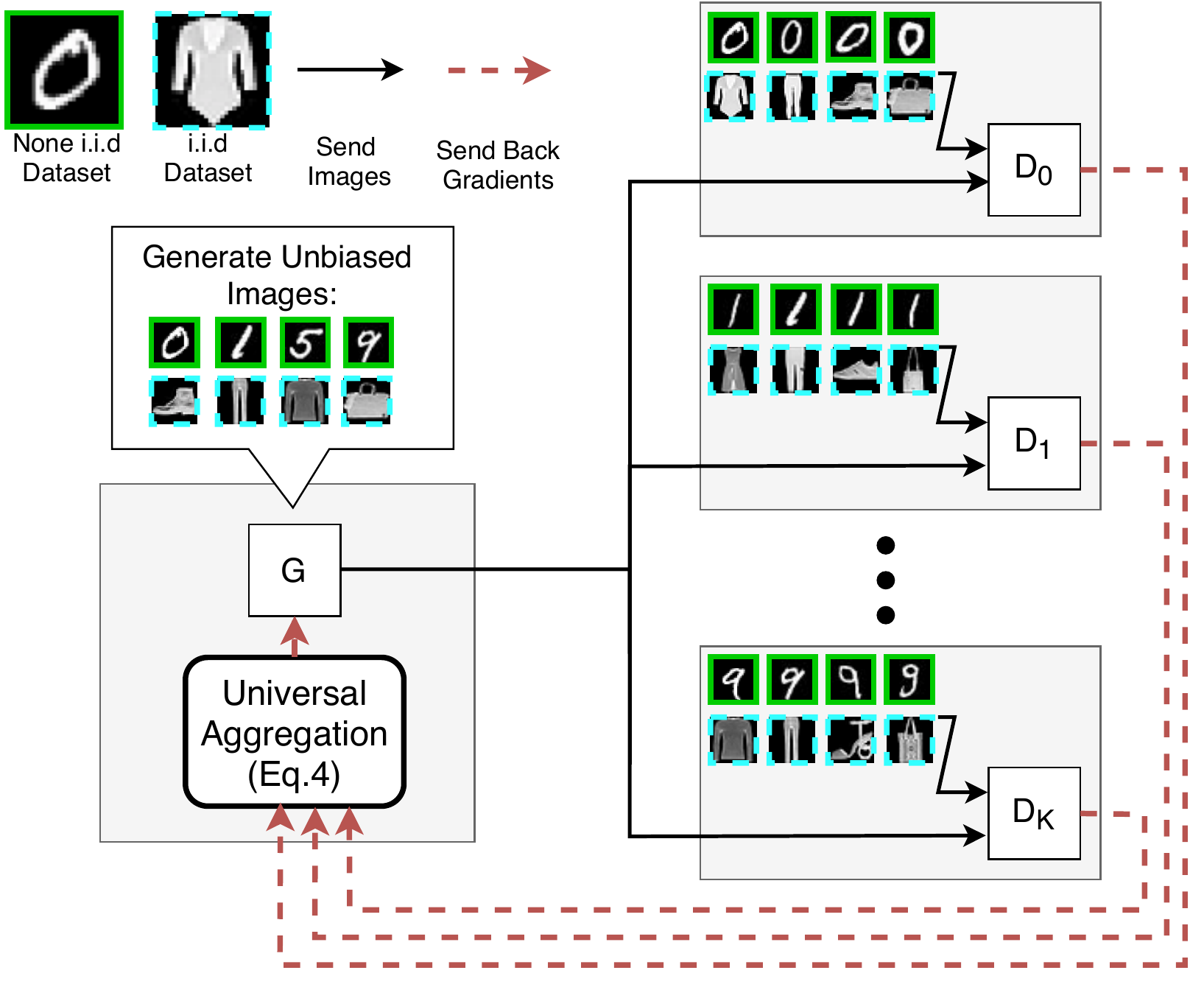}
    % \caption{In real life settings, the data centers always preserve private data with non-identical distribution(here we use MNIST and Fashion-MNIST as an example). Traditional aggregation method suffer from catastrophic bias: the generator only learn the common patterns in data and forget non-identical distribution data. Our UA-GAN is a generic framework for images with different type of distribution that applies to multi-discrminator GAN architecture.}'
    % \vspace{-.5em}
    \caption{\textbf{UA-GAN framework}: 
    The data from multiple entities may share some common distribution while retain  its own individual distribution.
   Here, we use the MNIST dataset that the 10 digits belong to different data center as common pattern, and MNIST Fashion dataset that randomly split into different data centers as distinct features. (see Sec.~\ref{sec_exp_real} for details). We can see the UA-GAN carefully aggregates the feedback from different entities and manage to digest the universal distribution generate unbiased synthetic data.
%   \vspace{-2em}
   }
%   does not suffer from any bias. 
    % Avg-GAN shows a strong bias toward the common pattern, i.e., the Fashion images. It forgets the distinct features for each site (classes 0 to 9 of MNIST, one for each site). 
    %  combined with non-identical and i.i.d images. 
    % \vspace{-.6em}
    \label{fig:arch1}
    % \vspace{-.6em}
% \end{figure}
% \end{minipage}
\end{figure}
% \vspace{-0.5in}
 In summary, the contributions are threefold.
    \begin{itemize} [leftmargin=5.5mm,noitemsep,topsep=0pt,parsep=0pt,partopsep=0pt]
        \item We propose UA-GAN, a novel federated GAN approach that aggregates feedback from local discriminators through their odds value rather than posterior probability.
        \item We prove that UA-GAN correctly learns the mixture of local distributions when they are perfectly modeled by local discriminators. 
        \item We prove when the discriminators are suboptimal in modeling their local distributions, the generator's approximation error is also linear. We also show that our bound is tight.
        % Corollary \ref{main_cor} shows if all discriminators $D_1,...,D_K$ behave nearly optimally, the target distribution of the generator using  UA framework is almost the same as real data distribution. In addition, we provide a lower bound to show the tightness of suboptimality.
    \end{itemize}

We show with various experiments that our method (UA-GAN) outperforms the state-of-the-art federated GAN approaches both qualitatively and quantitatively. 

Training on large scale heterogeneous datasets makes it possible to unleash the power of GANs. Federated GANs show their promise in utilizing unlimited amount of sensitive data without privacy and regulatory concerns. Our method, as the first theoretically guaranteed GAN, will be one step further in building such a foundation. Fig. \ref{fig:arch1} shows the workflow of UA-GAN.

\section{Related Work}
% \cc{This section is too long.}
\textbf{The Generative Adversarial Networks (GANs)} 
% The GANs \cite{goodfellow2014generative} are proposed to model data distributions. GANs are widely used in various machine learning and computer vision tasks \cite{zhang2018metagan, liu2019few, shaham2019singan, dai2017good, kumar2017semi}. %Different from traditional generative models which use kernel density functions or k-Nearest-Neighbours for density estimation, GANs achieve the distribution modeling by using a generator and a discriminator playing a game against each other. 
% The DCGAN \cite{} uses the convolutional neural networks in the generator and discriminator for image data. 
 have enjoyed much success in various machine learning and computer vision tasks \cite{zhang2018metagan, liu2019few, shaham2019singan, dai2017good, kumar2017semi}. 
Numerous methods are proposed for GAN training, such as Spectral Normalization (SN) \cite{miyato2018spectral}, zero-centered gradient penalty \cite{mescheder2018training, thanh2019improving}, WGAN \cite{arjovsky2017wasserstein}
, WGAN-GP \cite{gulrajani2017improved}, WGAN-TS \cite{liu2018two}, WGAN-QC \cite{Liu_2019_ICCV} etc. 
% \cite{salimans2016improved} proposed some tricks such as feature matching, batch discrimination, etc. to stabilize GAN training. are most widely used techniques for GAN training.  
A common approach in practice is the conditional GAN (cGAN)
\cite{mirza2014conditional}, which uses supervision from data (e.g., class labels) to improve GAN's performance.
% cGAN's idea is applied in state-of-the-art GANs like BigGAN \cite{brock2018large}, GauGAN \cite{park2019semantic} etc. for image generation. 
% Apart from the methodology part, recent progress on advanced network architectures \cite{karras2019style, Karras2019stylegan2, karnewar2019msg} also significantly improves GAN training, and facilitates GANs producing highly realistic images. 
% In paticular, this work is built upon GAN with Jensen-Shannon entropy loss. 
% Our approach aims to let
% GAN  learn a global distribution in a restricted setting where all discriminators only have domestic perspectives.

% , a multi-discriminator training framework is designed for a perturbed majority vote to obtain differential privacy guarantee. In \cite{ghosh2018multi}, a multi-generator GAN is proposed for learning multi-modes distributions. 
% \subsection{Multi-discriminator GAN}
% ,yoon2018pategan,ghosh2018multi
 \textbf{Multi-discriminator/-generator GANs} have been proposed for various learning tasks. To train these GANs, one common strategy is to directly exchange generator/discriminator model parameters during training \cite{xin2020private,hardy2019md}. This is very expensive in communication; a simple ResNet18 \cite{he2016deep} has 11 million parameters (40MB). 
 Closest to us is MD-GAN~\cite{hardy2019md}, which aggregates feedbacks (gradients) from local discriminators through averaging. It also swaps parameters between discriminators. None of these methods provide theoretical guarantee as ours. Meanwhile, our method is the only one without model swapping, and thus is much more efficient in bandwidth consumption. 
\textbf{Federated Learning (FL)} 
 \cite{kairouz2019advances,mcmahan2016communication}
offers the opportunity to integrate sensitive datasets from multiple sources through distributed training. Many works have been done tackling practical concerns in FL, such as convergence under Non-IID data assumption \cite{yu2019linear,lian2017can,Li2020On}, decentralized SGD without freezing parameters \cite{recht2011hogwild,nguyen2018sgd}, communication efficiency \cite{konevcny2016federated,li2019communication}, provable privacy guarantees \cite{alistarh2017qsgd,wei2020federated}. 
Federated GAN is also of great interest from a federated learning perspective. A successful federated GAN makes it possible to train a centralized model (e.g., a classifier) using data synthesized by the centralized generator. This becomes a solution when existing trained FL model needs to be replaced and updated by advanced machine learning approaches
, as one can retrain the model at any time using the generator.
It also alleviates some privacy concerns of FL, e.g., the gradient leakage problem~\cite{NIPS2019_9617}.

\section{Method}

To introduce our algorithm, we first introduce notations and formalize the mixture distribution learning problem. Next, we present our Universal Aggregation approach and prove that it is guaranteed to learn the target mixture distribution. We also analyze the suboptimality of the model when local discriminators are suboptimal. 
For ease of exposition, we mostly use ordinary GAN to illustrate the algorithm and prove its theoretical properties. At the end of this section, we extend the algorithm, as well as its theoretical guarantees, to conditional GAN (cGAN) \cite{mirza2014conditional}. The empirical results in this work are established on the cGAN since its training is much more controllable, thanks to the additional supervision by the auxiliary variable (e.g., classes of images). 
% Next we will introduce our multi-discriminator aggregation framework.

\myparagraph{Notations and problem formulation.}  We assume a cross-silo FL setting \cite{kairouz2019advances}, i.e., $K$ entities hosting $K$ private datasets $\mathcal{D}_1,...,\mathcal{D}_K$, with size $n_1, \cdots, n_K$. The total data size $n=\sum_{j=1}^K n_j$. The overall goal is to learn a target mixture distribution 
\begin{equation}
    p(x) = \sum\nolimits_{j=1}^K \pi_j p_j(x),
    \label{eq:target}
\end{equation}
in which a component distributions $p_j(x)$ is approximated by the empirical distribution from the $j$-th local dataset $\mathcal{D}_j$. 
The mixing weight $\pi_j$ is computed using the fraction of dataset $\mathcal{D}_j$: $\pi_j=n_j/n$. %\cc{Called $\pi_j$ the prior. is this ok?}
In general, different mixture components $p_i(x)$ and $p_j(x)$ may (but not necessarily) be non-identical, namely, $\exists x$, such that $p_i(x)\neq p_j(x)$. 

% share common support but behave dissimilarly. We call two distributions $p_i(x)$ and $p_j(x)$ non identical distribution if  $\exists (x) \in \supp(p_i(x)), \ p_j(x)/p_i(x) \neq 1$. We consider following mixture distribution learning problem: One wants to learn the global distribution which is a mixture of multiple components: $p(x)=\sum_{j} \pi_j p_j(x)$ where $\pi_j$s are the weight of each individual distribution with $\pi_j \geq 0$ ,$ \sum_{j=1}^{K} \pi_j =1$. In practice we compute $\pi_j$ by fraction of dataset $\mathcal{D}_j$: $\pi_j=n_j/n$ and approximate $p_j(x)$ using dataset $\mathcal{D}_j$ as its empirical distribution
	 
% 	 For simplicity  we can write $p(x)=\sum_{j} \pi_j(y) p_j(x)$  where $\pi_j(y)=[\#~\text{of} ~y~\text{in}~  \mathcal{D}_j ]/ n $.
% 	 %$\pi_j(y)=\text{[\# of y in }  \mathcal{D}_j ]/ \text{n} $. 
% 	 Throughout this paper we assume  dictionary of $y$ and  value of $\pi_j(y)$ is public known knowledge and will focus on the formulation of mixture distribution as  $p(x)=\sum_{j} \pi_j(y) p_j(x)$. 
% 	 For simplicity, we set=\pi_js_j(y)$ thus
% \cc{Cannot $s_j(y)$ be absorbed by $\pi_j(y)$?}
% \cc{This whole thing about conditional GAN is makeing things rather complicated. Can you first make the statement using unconditional GAN, and then extend it a conditional GAN?}

%  \begin{wrapfigure}{r}{0.8\textwidth}
%  	\begin{minipage}[t]{7.5cm}
\begin{algorithm}[t]
\caption{Training Algorithm of UA-GAN.}
 \begin{algorithmic}[1]\label{alg:UA-GAN}
%  \label{alg:AGF}
\STATE {\bfseries Input:} Batch size $m$, datasets $\{\mathcal{D}_j\}$, size of datasets $\{\pi_j=\frac{n_j}{n}\}$. 
\STATE {\bfseries Output:} $G$, $D_j, \forall j \in [K]$.
  \FOR{$t=1,\cdots,T$}
%   \STATE{-- Sample $y_i, i \in [m]$ according to  $supp(y)$ and  $\pi_j(y), j\in[K]$.}
\STATE \COMMENT{Work at the central server.}

  \STATE{$G$ generates synthetic data: $\hat{x}_i = G(z_i)$, $i=1,\cdots,m$. }
  \STATE{Send batch of synthetic data $\mathcal{D}_{syn}=\{\hat{x}_1,\cdots,\hat{x}_m\}$ to all $K$ sites.}
  \FOR{$j = 1, \cdots, K$}
  \STATE \COMMENT{Work at each local site.}
  \STATE{Update the local discriminator, $D_j$, using real samples from $ \mathcal{D}_j$ and synthetic data batch, $\mathcal{D}_{syn}$, based on Eq.~\ref{Dis}.}
\STATE{Output predictions and gradients for synthetic data $D_j(\hat{x}_i)$, $\partial D_j(\hat{x}_i) / \partial \hat{x}_i$, $i =1, \cdots,m$. Send them to the central server.}   \ENDFOR
  \STATE \COMMENT{Work at the central server.}
  \STATE{Simulate value of $D_{ua}(\hat{x}_i)$ via Eq.~\ref{aggre}, $\forall i$.}
  \STATE{Update $G$ based on Eq.~\ref{Gen}, using gradients from $D_j$'s.}
%   \STATE{-- GeberSample $m$ noises $\{z_1,...,z_m\}$, and $m$ labels $\{ \hat{y}_1, ..., \hat{y}_m\}$ }
%   \STATE{-- Let $\hat{x}_i = G(z_i, \hat{y}_i), \forall i \in [m]$. Send $(\hat{x}_i, \hat{y}_i),\forall i \in [m]$ to all $K$ nodes.}
%   \STATE{-- Each discriminator outputs $D_j(\hat{x}_i, \hat{y}_i), \forall i \in [m]$ and send to generator.}
%   \STATE{-- The aggregated discriminator output for each ($\hat{x}_i, \hat{y}_i$) is
%   \[D(\hat{x}_i|\hat{y}_i)=\frac{\sum_{j}^K  \pi_j (\hat{y}_i) \frac{D_j(\hat{x}_i|\hat{y}_i)}{1-D_j(\hat{x}_i|\hat{y}_i)}}{1+\sum_{j}^K \pi_j (\hat{y}_i) \frac{D_j(\hat{x}_i|\hat{y}_i)}{1-D_j(\hat{x}_i|\hat{y}_i)}}
%   \]
%   }
%   \STATE{-- Update $G$ by descending its stochastic gradient:
%   \[ \nabla_{\Phi_G} \frac{1}{m} \sum_{i=1}^m
%   \log (1-D(\hat{x}_i|\hat{y}_i)).\]}
  \ENDFOR
%   \\The gradient-based updates can use any standard gradient-based learning rule.
 \end{algorithmic}

\end{algorithm}
%  \vspace{-.05in}
% \end{minipage}%
% \end{wrapfigure}
% The UA approach GAN jointly trains a centralized generator and multiple private discriminators hosted at different sites. 
\myparagraph{Universal Aggregation GAN:} Now we are ready to introduce our multi-discriminator aggregation framework.  A pseudo-code of UA framework can be found in Algorithm \ref{alg:UA-GAN}. We have a centralized (conditional) generator $G(z)$ seeking to learn the global distribution $p(x)$. 
In each local site, a discriminator $D_j(x)$ has access to local dataset $\mathcal{D}_j$. 
Note data in $\mathcal{D}_j$ are only sampled from $p_j(x)$.
During training, the generator generates a batch of synthetic data to all $K$ sites.
 	The $j$-th discriminator seeks to minimize the cross entropy loss  of GAN from a local perspective~\cite{goodfellow2014generative}:
%  	\vspace{-.07em}
	\begin{equation}
	\begin{aligned}
% 	\vspace{-.03em}
	&\max\limits_{D_j}V(G,D_j) =\mathbb{E}_{x\sim p_{j}(x)} [\log D_j(x)] +\mathbb{E}_{z\sim \mathcal{N}(0,I_d)} [\log(1-D_{j}(G(z)))]
	%	&=\sum \pi_j\int\limits_{y} s_j(x)\int\limits_{x} p_j(x)log D_j(y,x)+q(x)log(1-D_j(y,x)) dxdy
% 	 \vspace{-.07em}
	\end{aligned}
	\label{Dis}
	\end{equation}
% 	\vspace{-.07em}%
% 	Similar to the definition in \cite{goodfellow2014generative}, we say a discriminator $D_j$ is optimal if it minimizes Eq. (\ref{Dis}).
	 To formalize the generator training, we first introduce odds value. It is an essential quantity for our algorithm and its analysis.  
	\begin{definition}[odds value]
	Given a probability $\phi \in (0,1)$, its odds value is $\Phi(\phi ) \triangleq \frac{\phi}{1-\phi}$. Note the definition requires $\phi\neq 1$. 
    \end{definition}
    % \vspace{-.1em}
    Also it is straightforward to see $\phi  = \frac{\Phi(\phi)}{1+\Phi(\phi)}$.

    % 	by collection score of different $D_j(x)$s. 
	The central idea of UA Framework is to simulate a centralized discriminator $D_{ua}(x)$ which behaves like the mixture of all local discriminators (in terms of odds value). A well behaved $D_{ua}(x)$ can then train the centralized generator $G$ using its gradient, just like in a classical GAN.
	
We design $D_{ua}$ so that its odds value $\Phi(D_{ua}(x))$ is identical to the mixture of the odds values of local discriminators:
    % \vspace{-.02em}
    \begin{equation} \label{aggre_2}
        \Phi(D_{ua}(x))=\sum\nolimits_{j=1}^K  \pi_j \Phi(D_j(x)).
    \end{equation} 
Given $\Phi(D_{ua}(x))$, we can compute $D_{ua}(x)$ as
% 	\vspace{-.07em}
    \begin{equation} \label{aggre}
        D_{ua}(x)=\frac{ \Phi(D_{ua}(x))}{1+ \Phi(D_{ua}(x))}
    \end{equation} 

	Once the central discriminator $D_{ua}$ is simulated, the generator can be computed by minimizing the generator loss :  
% 	\vspace{-.1em}
	\begin{equation} \label{Gen}
	\begin{aligned}
	\min\limits_{G}V(G,D_{ua})&= \mathbb{E}_{x\sim p(x)} [\log D_{ua}(x)] +\mathbb{E}_{z\sim \mathcal{N}(0,I_d)} [\log(1-D_{ua}(G(z))]
	\end{aligned}
	\end{equation}
Note that mathematically, Eq.~(\ref{Gen}) can be directly written in terms of local discriminators $D_j$'s (by substituting in Eqs (\ref{aggre_2}) and (\ref{aggre})). In implementation, the simulated central discriminator can be written as a pytorch or tensorflow layer.  

\myparagraph{Intuition.} 
The reason we define $D_{ua}$'s behavior using a mixture of odds values instead of a mixture of the predictions is mathematical. It has been shown in ~\cite{goodfellow2014generative} that a perfect discriminator learning a data distribution $p(x)$ and a fixed generator distribution $q(x)$ satisfies $D(x)=\frac{p(x)}{p(x)+q(x)}$. It can be shown that only with the odds value equivalency, this optimal solution of the central discriminator $D(x)$ can be recovered if each individual discriminator is optimal, i.e., $D_j(x)=\frac{p_{j}(x)}{p_{j}(x)+q(x)}$.   This is not true if we define the central discriminator behavior using the average prediction, i.e., $D_{ua} = \sum_j \pi_j D_j$. More details can be found in Theorem (\ref{Unbiased}) and its proof.

    	\begin{remark}[Privacy Safety]
    	For federated learning, it is essential to ensure information of real data are not leaked outside the local site. This privacy safety is guaranteed in our method. To optimize $G$ w.r.t.~Eq.~(\ref{Gen}), we only need to optimize the second term and use gradient on synthetic images $G(z)$ from local discriminators. 
    	
% 	In the UA framework, discriminators and generator deal with distinct loss function which is different from existing methods. In order to update the generator, the score $D_j(G(z))$ with corresponding gradient on each synthetic image needs to be transmitted from each discriminator to the generator  and then aggregated for update. Note that the score on real images $D_j(x)$ does not need to be transmitted since the central generator doesn't require such information for update. 
	\end{remark} % shaonian, gai guo de comments dian yi xia resolve
	
% 	    	\begin{remark}[Privacy Safety]
% 	In the UA framework, discriminators and generator deal with distinct loss function which is different from existing methods. In order to update the generator, the score $D_j(G(z))$ with corresponding gradient on each synthetic image needs to be transmitted from each discriminator to the generator  and then aggregated for update. Note that the score on real images $D_j(x)$ does not need to be transmitted since the central generator doesn't require such information for update. 
% 	\end{remark} % shaonian, gai guo de comments dian yi xia resolve

One important concern is about the optimal discriminator condition. 
   $D_{ua}(x)$ is designed to be optimal only when $D_j$'s are optimal. We need to investigate the consequence if the local discriminators $D_j$'s are suboptimal. We will provide an error bound of the learned distribution w.r.t., the suboptimality of $D_j$'s in Corollary (\ref{main_cor}).

	\subsection{Theoretical Analysis of UA-GAN}\label{ana_sec}
% 	 and with \emph{ suboptimal discriminator} assumptions. 
In this section, we prove the theoretical guarantees of UA-GAN. First, we prove the correctness of the algorithm. We show that if all local discriminators can perfectly learn from their local data, the algorithm is guaranteed to recover the target distribution (Eq.~(\ref{eq:target})). Second, we discuss the quality of the learned generator distribution when the local discriminators are suboptimal, due to real-world constraints, e.g., insufficient local data. We show that the error of the learned distribution is linear to the suboptimality of the local discriminators. All results in this section are for original (unconditional) GANs. But they can be extended to conditional GANs (see Sec.~(\ref{sec:cgan})). Due to space constraints, we only state the major theorems and leave their proofs to the supplemental material.

	\subsubsection*{Correctness of UA-GAN}	\label{sec_optimal_dis}
% 	Theorem \eqref{fail} gives a failure case of simple average scheme: optimal discriminator can not lead to correct target distribution. 
% 		\begin{lemma}\label{lem1}
% 		When generator $G$ is fixed,  the optimal discriminator $D_j(y,x)$ is :\\
% 		\begin{equation}
% 		D_j(y,x)=\frac{p_j(x)}{p_j(x)+q(x)}
% 		\end{equation}
% 	\end{lemma}
The correctness theorem assumes all local discriminators behave optimally. 
\begin{assume}[Optimal Local Discriminator]
    Assume all local discriminators are \emph{optimal}, i.e., they learn to predict whether a data is true/fake perfectly. Let $q(x)$ be the probability of the current generator $G$. A local discriminator is optimal iff $D_j(x) = \frac{p_j(x)}{q(x) + p_j(x)}$.
\label{asm:local-optimal}
\end{assume}
	Theorem (\ref{Unbiased}) states the unbiasedness of UA-GAN: with optimal local discriminators, the generator learns the target distribution. 
% 	The theorem shares a similar spirit with Theorem 1 in 
	\begin{thm} [Correctness] \label{Unbiased}
	    Suppose all discriminators $D_j$'s are optimal. $D_{ua}(x)$ is computed via Eq.~(\ref{aggre_2}).  
        Denote by $q$ the density function of data generated by $G$.
	    Let $q^*(\cdot)$ be the optimal distribution w.r.t.~the Jenson Shannon divergence loss  :
	   % \cc{What loss? Jenson Shannon divergence? Should you explain more here? Or just assume ppl follows as it is folklore for GAN?}
	   % \vspace{-.1em}
	    \begin{equation}
	    q^*:= \argmin\limits_{q}L(q)= \mathbb{E}_{x\sim p(x)} [\log D_{ua}(x)] +\mathbb{E}_{x\sim q(x)} [\log(1-D_{ua}(x)].
	    \label{eq:jenson-shannon}
	    \end{equation}
        % \vspace{-.1em}
	    We have $q^\ast$ equals to the true distribution, formally, $q^* = p$.
	\end{thm}
	The proof mainly establishes that when $D_j$'s are optimal, $D_{ua}$ is also optimal. With the optimality of $D_{ua}$ proved, the rest follows the correctness proof of the classic GAN (Theorem 1 in \cite{goodfellow2014generative}). More details are in the supplemental material.

	\subsubsection*{Analysis of the Suboptimal Setting} \label{sec_sub_opt}
	
	In centralized learning setting, an optimal discriminator is a reasonable assumption since the model has access to all (hopefully sufficient) data. However, in federated GAN setting, available data in some site $\mathcal{D}_j$ may be limited. One natural question in this limited data scenario is: how would the generator behave if some local discriminators are suboptimal? We address this theoretical question.
	
We first focus on a single discriminator case. We show the behavior of a perturbed version of  Jensen-Shannon divergence loss \cite{guha2007sketching,lin1991divergence,csiszar2004information}. The suboptimality of a central discriminator $D(x)$ is measured by the deviation in terms of the odds value. Denote by $q(x)$ the generator distribution of the current $G$. Ideally, the odds value of an optimal discriminator should be $p(x)/q(x)$. We show that a suboptimal $D$ with $\delta$ deviation from the ideal odds value will result in $O(\delta)$ suboptimality in the target distribution.
    % $\frac{D(x)}{1-D(x)}=\frac{p(x)}{q(x)}$
	\begin{thm}[Suboptimality Bound for a Single Discriminator] \label{main}
		Suppose a discriminator $\widetilde{D}(x)$ is a perturbed version of the optimal discriminator $D(x)$, s.t.~$\Phi(\widetilde{D}(x))= \Phi(D(x)) \xi(x)$ with  $|1-\xi(x)| \leq \delta$ and $\delta \leq 1/8$. 
		Let $q^\ast$ be the optimal distribution of the Jensen-Shannon divergence loss based on the perturbed discriminator 
		\begin{equation}
	    q^\ast: = \argmin\limits_{q}L(q)= \mathbb{E}_{x\sim p(x)} [\log \widetilde{D}(x)] +\mathbb{E}_{x\sim q(x)} [\log(1-\widetilde{D}(x)].
	    \label{eq:jenson-shannon-2}
	    \end{equation}
	    Then $q^\ast$ satisfies
	 $|q^\ast(x)/p(x)-1| \leq 16 \delta, \forall x$.
	\end{thm}

   This theorem shows that the ratio of the learned distribution $q^\ast$ is close to the target true distribution $p$ when the suboptimality of $D_j$ is small. To the best of our knowledge, this is the first bound on the consistency of Jensen-Shannon divergence with suboptimal discriminator, even for a classical GAN.  
    % One concern about above analysis is that the  odds value is not an intuitive quantity compared  to simple metrics e.g., $|D(x)-\tilde{D}(x))|$ and $|D(x)/ \tilde{D}(x)-1|$. It comes to the question   whether such bound is informative. 
    
Next, we show that the bound is also tight.   

	  \begin{thm}[Tightness of the Bound in Theorem (\ref{main})] \label{tightness}
	  Given a perturbed discriminator $\widetilde{D}(x)$ of the optimal one $D(x)$, s.t.~$\Phi(\widetilde{D}(x))= \Phi(D(x))\xi(x)$ 
% 		Suppose a discriminator $\tilde{D}(x)$ is a perturbed version of discriminator $D(x)$ s.t. $\frac{D(x)}{1-D(x)}=\frac{p(x)}{q(x)}$ is perturbed by $\xi(x)$ 
		with  $|\xi(x)-1| \geq \gamma$ and $\gamma \leq 1/8$. The optimal distribution $q^\ast$ as in Eq.~(\ref{eq:jenson-shannon-2}) satisfies $|q^\ast(x)/p(x)-1| \geq \gamma/16$, $\forall x$.
    \end{thm}   
	 
	 Next we extend these bounds for a single discriminator to our multi-discriminator setting. This is based on Theorem (\ref{main}) and the linear relationship between the local discriminators and the central discriminator.
	 \begin{cor} [Suboptimality Bound for UA-GAN] \label{main_cor}
	 	Assume suboptimal local discriminators $\widetilde{D_j}(x)$ are the perturbed versions of the optimal ones $D_j(x)$. And the suboptimality is bounded as: $\Phi(\widetilde{D_j}(x))=\Phi(D_j(x))\xi_j(x)$  with  $|\xi_j(x)-1| \leq \delta \leq 1/8$, $\forall x$. The centralized discriminator $\widetilde{D_{ua}}(x)$ is computed using these perturbed local discriminators such that $\Phi(\widetilde{D_{ua}}(x))=\sum_{j=1}^K\pi_j\Phi(\widetilde{D_j}(x))$. 
% 	 	Let $q(x)$ be the density (mass) function of $G(z)$. 
		Let $q^\ast$ be the optimal distribution of the Jensen-Shannon divergence loss based on the perturbed UA discriminator $\widetilde{D_{ua}}$ 
		\begin{equation}
	    q^\ast: = \argmin\limits_{q}L(q)= \mathbb{E}_{x\sim p(x)} [\log \widetilde{D_{ua}}(x)] +\mathbb{E}_{x\sim q(x)} [\log(1-\widetilde{D_{ua}}(x)].
	    \label{eq:jenson-shannon-3}
	    \end{equation}
	    Then $q^\ast$ satisfies $|q^\ast(x)/p(x)-1|=O(\delta)$.   In particular, the optimal distribution $q^*(x)$ has $O(\delta)$ total variation distance to the target distribution $p(x)$.
	 	
	 \end{cor}
	 
	 Note that the lowerbound of the suboptimality for single discriminator (Theorem (\ref{tightness})) can also be extended to UA-GAN similarly. 

	 \begin{remark}
	    The consistency gap in  Corollary (\ref{main_cor}) assumes a uniform suboptimality bounded for all local discriminators. In practice, such assumption may not be informative if the sizes of $\mathcal{D}_j$'s data are highly imbalanced. It would be interesting to relax such assumption and investigate the guarantees of UA-GAN w.r.t.~the expected suboptimality of $D_j$'s. 
	 \end{remark}
% 	 \begin{remark}
% 	    Strikingly, the well know model collapse problem \cite{arora2017generalization,arora2017gans} does not appear in our analysis. When model collapse happens, the mass of $q(x)$ will be concentrated inside a small subset of  $support(p(x))$ thus a large fraction of $x$ in $support(p(x))$ will have zero mass.  We accidentally avoid such troublesome situation via the odds value quantity. The odds value of an optimal discriminator is  $p(x)/q(x)$, which implicitly requires $q(x)$ to be nonzero for any $x$ in the support of $p(x)$. Such observation suggests a potential strategy to handle the model collapse case by enforcing the generator to be uniformly activated within domain of target distribution $p(x)$. 
% % \cc{This is incorrect with all the new notations. Plz fix accordingly.}
%     \end{remark}
	   % The consistency gap guarantee in  Corollary \eqref{main_cor} is mainly based the assumption that the odds value perturbation is uniformly bounded..
% 	 \end{remark}
% \vspace{-.3em}
\subsection{Universal Aggregation Framework for Conditional GAN}
\label{sec:cgan}
% \vspace{-.1em}

Our algorithm and analysis on unconditional GANs can be generalized to the more practical Conditional GAN~\cite{mirza2014conditional}. 
A conditional GAN learns the joint distribution of $p(x,y)$. Here $x$ represents an image or a vectorized data, and $y$ is an auxiliary variable to control the mode of generated data (e.g., the class label of an image/data). 
Conditional GAN is much better to train in practice and is the common choice in most existing works.
This is indeed the setting we use in our experiments.

The target distribution of the learning problem becomes a joint mixture distribution: 
$$p(x,y)=\sum_{j}\pi_j \omega_j(y) p_j(x,y),$$ 
in which $\pi_j = n_j/n$ and $\omega_j(y)$ is the proportion of class $y$ data within the $j$-th local dataset $\mathcal{D}_j$.
%  The mixture distribution can be formulated as $p(x,y)=\omega_j(y) p_j(x|y),  j\in[K]$ where $\omega_j(y)=\pi_j\cdot[\text{\# of y in }\mathcal{D}_j].$ 
 We assume $\pi_j$, and the dictionary of $y$ and its fractions in each $\mathcal{D}_j$, $\omega_j(y)$ are known to the public. In practice, such knowledge will be used for generating $y$. Formally, $y \sim \sum_{j=1}^{K}\pi_j\omega_j(y)$. 
 
 To design the UA-GAN for the conditional GAN, the odds value aggregation in formula Eq.~(\ref{aggre_2}) needs to be adapted to:
    \begin{equation*}
    % \vspace{-.1em}
         \Phi(D_{ua}(x|y))=\sum_{j=1}^K  \pi_j\omega_j(y) \Phi(D_j(x|y)).
    %  \vspace{-.1em}
    \end{equation*}
    % $$$$
    % \vspace{-.1em}
    The computation of $D_{ua}(x|y)$ and the update of $G$ and $D_j$'s need to be adjusted accordingly. The theoretical guarantees for unconditional GANs can also be established for a conditional GAN. Due to space limitation, we leave details to the supplemental material.

\section{Experiments} 
\label{sec_exp}
% \vspace{-.7em}
On synthetic and real-world datasets, we verify that UA-GAN can learn the target distribution from both i.i.d and non-identical local datasets. We focus on conditional GAN~\cite{mirza2014conditional} setting as it is the common choice in practice.

\begin{figure}[]
    \centering
    \begin{tabular}{ccc}
    \textbf{UA-GAN} & \includegraphics[align=c,width=0.75\textwidth]{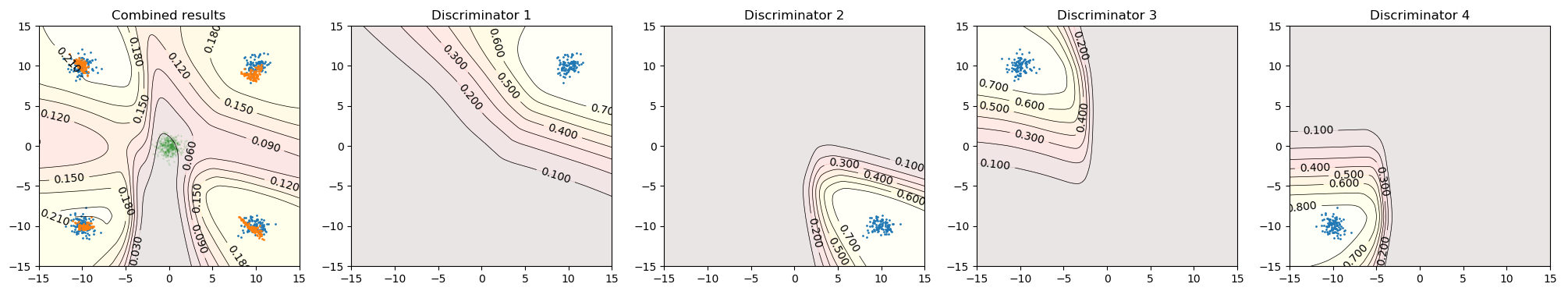}\\
    \textbf{Avg-GAN} & \includegraphics[align=c,width=0.75\textwidth]{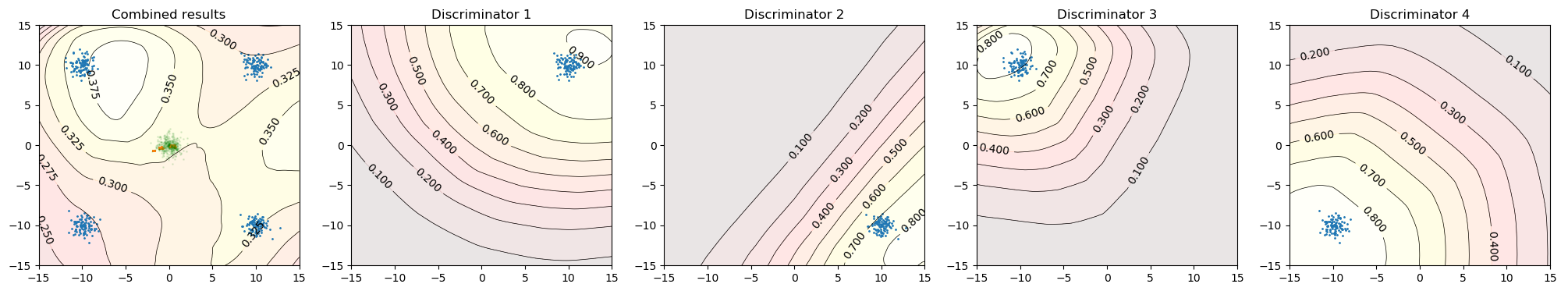}\\
    \textbf{MD-GAN} & \includegraphics[align=c,width=0.75\textwidth]{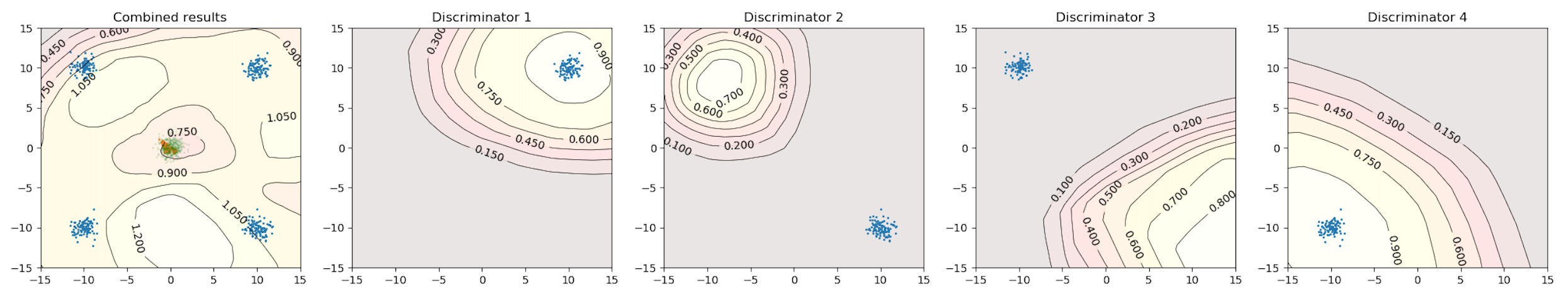}
    \end{tabular}
    \caption{Results on a toy dataset by UA-GAN, Avg-GAN and MD-GAN. UA-GAN can learn four Gaussians, whereas Avg-GAN and MD-GAN fail.}
    \label{fig:toy}
    % \vspace{-0.7em}
\end{figure}

% \begin{figure}[t]
% 	\vspace{-.3in}
%     \centering
%     {\scriptsize \textbf{UA-GAN}}\\
%         \includegraphics[width=0.8\textwidth]{pics/toy-ours.png}\\
%     {\scriptsize \textbf{Avg-GAN}}\\
%     \includegraphics[width=0.8\textwidth]{pics/toy-average.png} \\
%     {\scriptsize \textbf{MD-GAN}}\\
%     \includegraphics[width=0.8\textwidth]{pics/MDGAN_toy3.png}
%     \caption{Results on a toy dataset by UA-GAN, Avg-GAN and MD-GAN. UA-GAN can learn four Gaussians, while Avg-GAN and MD-GAN fails.}
%     \label{fig:toy}
%     \vspace{-.1in}
% \end{figure}

% \vspace{-.2em}
\subsection{Synthetic Experiment}
% \vspace{-.1em}
We evaluate UA-GAN on a toy dataset.
See Fig.~\ref{fig:toy} first row.
The toy dataset has 4 datasets, generated by 4 Gaussians centered at (10, 10), (10,-10), (-10,10), (-10,-10) with variance 0.5. Data samples are shown as blue points. The central generator takes Gaussian noise centered at $(0, 0)$ with variance of 0.5 (green points) and learns to transform them into points matching the mixture distribution (orange points). 
The first figure shows the generator successfully recovers the Gaussian mixture. The contours show the  central discriminator $D_{ua}$ calculated according to  Eq.~\ref{aggre_2}.  The generated points (orange) are evenly distributed near the 4 local Gaussians' centers. The 2nd to 5th figures show the prediction of the local discriminators ($D_j$'s), overlaid with their respective samples $\mathcal{D}_j$ in blue. 
% \cc{Remove orange points in columns 2 to 5.}

Meanwhile, we show in the second row the results of the average scheme, called Avg-GAN, which averages local discriminators' outputs to train the generator.
The first figure shows that the learned distribution $D_{avg} = \frac{1}{K}\sum_j D_j$ is almost flat. The generated samples (orange) collapse near the origin. Although each individual discriminator can provide valid gradient information (see 2nd to 5th figures for $D_j$'s), naively averaging their outputs cannot achieve the target distribution, when the distributions are symmetric. We show the results of the MD-GAN by \cite{hardy2019md} in the third row. MD-GAN also adopts the average scheme, but randomly shuffle discriminators parameters during training. Similar to Avg-GAN, MD-GAN cannot learn the four Gaussians.

% 1. Toy sample. Show aggregated gradient works
% 2. Semi-sythetic data
% Given a well trained generator(s), we generate variants of data and distribute data into multiple entities. We train our distributed gan only via the sythetic data from multiple entities. We aim to see how is the performance changing with increasing number of samples.

\begin{table}[t]
	\begin{center}
		\begin{tabular}{l|ccc|ccc}
		    \midrule
		    \multirow{2}{*}{Dataset} 
		    & \multicolumn{3}{c}{non-identical Mnist + Fashion}
		    & \multicolumn{3}{c}{non-identical Mnist + Font} \\ 
		    \cmidrule{2-7}
		    & Accuracy$\uparrow$ & IS$\uparrow$ & FID$\downarrow$ & Accuracy$\uparrow$ & IS$\uparrow$ & FID$\downarrow$ \\ 
		    \midrule
		    
			Real & 0.943 & 3.620 $\pm$ 0.021 & 0
			& 0.994 & 2.323 $\pm$ 0.011 & 0 \\ 

			 Centralized GAN & 0.904  & 3.437 $\pm$ 0.021 & 8.35 & 0.979 & 1.978 $\pm$ 0.009 & 17.62 \\
			\midrule
			Avg GAN & 0.421 & 4.237 $\pm$ 0.023 & 72.80 & 0.822 & 1.517 $\pm$ 0.004 & 85.81 \\
			
			MD-GAN & 0.349  & 2.883 $\pm$ 0.020 & 102.00   & 0.480 & \textbf{2.090 $\pm$ 0.007} & 69.63   \\

			UA-GAN & \textbf{0.883} & \textbf{3.606 $\pm$ 0.020} & \textbf{24.60} & \textbf{0.963} & 1.839 $\pm$ 0.006 & \textbf{24.73}\\
			\bottomrule
		\end{tabular}
	\end{center}
% Best method among the three federated GANs is highlighted.
	\caption{Quantitative results on non-identical mixture datasets.  UA-GAN achieves better result compared with the Avg GAN and MD-GAN's aggregation method. }
% 	\vspace{-0.1in}
	\label{tab:acc}
% 	\vspace{-1em}
\end{table}
% \begin{table}[t]
% 	\begin{center}
% 	\vspace{-.1em}
% 		\begin{tabular}{lcccccc}
% 		    \midrule
% 		    \multirow{2}{*}{Dataset} 
% 		    & \multicolumn{3}{c}{i.i.d Mnist + Fashion}
% 		    & \multicolumn{3}{c}{i.i.d Mnist + Font} \\ 
% 		    \cmidrule{2-7}
% 		    & Accuracy$\uparrow$ & IS$\uparrow$ & FID$\downarrow$ & Accuracy$\uparrow$ & IS$\uparrow$ & FID$\downarrow$ \\ 
% 		    \midrule
		    
% 			Real & 0.943 & 3.620 $\pm$ 0.021 & 0
% 			& 0.994 & 2.456 $\pm$ 0.009 & 0 \\ 
% 			\midrule
% 			Regular GAN & 0.904  & 3.437 $\pm$ 0.021 & 8.35 & 0.983 & 2.257 $\pm$ 0.011 & 6.07 \\
			
% 			Avg GAN & 0.905 & 3.371 $\pm$ 0.026 & 12.83 & 0.977 & 2.221 $\pm$ 0.010  & 9.10 \\
			
% 			MD-GAN &0.904  & 3.346 $\pm$ 0.020 & 12.73 &0.979  & 2.192 $\pm$ 0.005 & 9.77   \\
% 			\midrule
% 			\textbf{UA-GAN} & 0.908 & 3.462 $\pm$ 0.024
%  & 11.82 & 0.975 & 2.201 $\pm$ 0.014 & 9.79\\
% 			\bottomrule
% 		\end{tabular}
% 			\vspace{-.05in}
% 	\end{center}
% 	\caption{The classification accuracies and IS, FID scores on two i.i.d mixture datasets. All of the three architecture could learn the right distribution with i.i.d datasets. \cc{Rename to Centralized GAN.}}	
% %	\vspace{-0.1in}
% 	\label{tab:acc_iid}
% 	\vspace{-.1em}
% \end{table}
% \vspace{-.2em}
\subsection{Real-World Mixture Datasets}
% \vspace{-.1em}
\label{sec_exp_real}
We evaluate our method on several mixture datasets, both i.i.d and non-identical. 

\myparagraph{Datasets.}
Three real-world datasets are utilized to construct the mixture datasets: MNIST~\cite{lecun1998gradient}, Fashion-MNIST~\cite{xiao2017fashion}, and Font dataset. 
% MNIST and Fashion-MNIST both have 10 classes of $28\times28$ grayscale images, with 60k samples for training and 10k for validation. 
We create the Font dataset from 2500+ fonts of digits taken from the Google Fonts database \cite{mo2002internet}.
Similar to MNIST, it consists of 10 classes of $28\times28$ grayscale images, with 60k samples for training and 29k samples for test. To make the differences more clear between font and handwrite images, we highlight the Font images with a circle when build the dataset. 
Using these foundation datasets, we create 2 different mixture datasets with non-identical local datasets: (1) non-identical MNIST+Fashion; (2) non-identical MNIST+Font. We uniformly sample Fashion/Font data for all 10 distributed sites. These are common patterns across all sites. Meanwhile, for each individual site, we add MNIST data with a distinct class among $0\sim9$. These data are distinguishable features for different sites. Ideally, a federated GAN should be able to learn both the common patterns and the site-specific data. Please see supplemental material for more details and samples from these mixture datasets. 

\myparagraph{Baselines.} 
We compare UA-GAN with Avg-GAN and another SOTA federated GAN, MD-GAN~\cite{hardy2019md}. We also include two additional baselines: \textbf{Centralized GAN} trains using the classic setting, i.e., one centralized discriminator that has access to all local datasets. Comparing with this baseline shows how much information we lose in the distributed training setting.
Another baseline is \textbf{Real}, which essentially uses real data from the mixture datasets for evaluation. This is the upper bound of all GAN methods. More details can be found in supplementary material.

\myparagraph{Evaluation metrics.}
We adopt \textbf{Inception score (IS)}~\cite{salimans2016improved}, \textbf{Frechet Inception distance (FID)}~\cite{heusel2017gans} to measure the quality of the generated images. 
% as ad-hoc measured correlated with the quality of generated images \cc{Do not understand this sentence}. 
As GAN is often used for training downstream classifiers, we also evaluate the methods by training a classifier using the generated data and report the \textbf{Classification Accuracy} as an evaluation metric.
This is indeed very useful in federated learning; a centralized classifier can be trained using data generated by federated GANs without seeing the real data from private sites. 
% The classification Accuracy is also critical to evaluate the model adaptivity problem in \cite{ourpaper}.
% one may need to leverage on the  
% ultimately the utility of these models is their performance 

% \footnote{If space is limited, remove this and next paragraphs}Inception score(IS) measures the quality of a generated image by computing the KL-divergence between the response produced by this image and the marginal distribution, i.e., the average response of all the generated images, using an Inception network \cite{szegedy2015going} trained on ImageNet. 

% \begin{equation}
%     IS(G) = \exp(\mathbb{E}_{x \sim p_y}[D_{KL}(p(y|x) || p(y))])
% \end{equation}
% Where $x$ is a generated image sampled from the learned generator distribution $p_g$, $\mathbb{E}$ is the expectation over the set of generated images, $D_{KL}$ is the KL-divergence between the conditional class distribution p(y|x) and the marginal class distribution. This score is limited to measuring the diversity of generated images. 

%FID compares the distributions of Inception embeddings of real ($p_r(x)$) and generated ($p_g(x)$ images. The distance measure is defined between two Gaussian distributions are:

% \begin{equation}
%     d^2((m_r, C_r),(m_g, C_g)) = ||m_r - m_g||^2 + Tr(C_r + C_g - 2(C_rC_g)^{\frac{1}{2}})
% \end{equation}
% where $(m_r, C_r), (m_g, C_g)$ denote the mean and co-variance of the real and generated image distributions respectively. 

\textbf{Discussion.}
The quantitative results on the two non-identical mixture datasets are shown in Table \ref{tab:acc}. UA-GAN significantly outperforms the other two federated GANs, Avg-GAN and MD-GAN. Its performance is even close to the centralized GAN, showing that our algorithm manages to mitigate the challenge of distributed training to a satisfying degree. 

\begin{figure}[]
    \centering
    \begin{minipage}{0.16\linewidth}
    \centering\includegraphics[width=\textwidth]{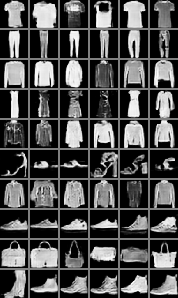} \\ (a) Avg GAN
    \end{minipage}
    \begin{minipage}{0.16\linewidth}
    \centering\includegraphics[width=\textwidth]{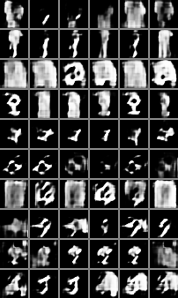}  \\ (b) MD-GAN
    \end{minipage}
    \begin{minipage}{0.16\linewidth}
    \centering\includegraphics[width=\textwidth]{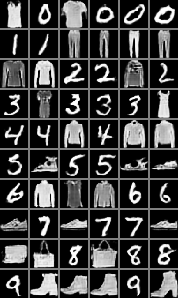} \\ (c)  UA-GAN
    \end{minipage}
    \begin{minipage}{0.16\linewidth}
    \centering\includegraphics[width=\textwidth]{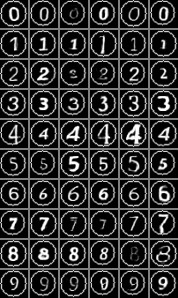}  \\ (d) Avg GAN
    \end{minipage}
    \begin{minipage}{0.16\linewidth}
    \centering\includegraphics[width=\textwidth]{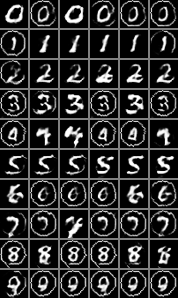}  \\ (e) MD-GAN
    \end{minipage}
    \begin{minipage}{0.16\linewidth}
    \centering\includegraphics[width=\textwidth]{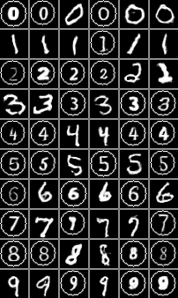} \\ (f) UA-GAN
    \end{minipage}
    \caption{Synthetic images on the non-identical MNIST+Fashion ((a),(b),(c)) and MNIST+font datasets ((d),(e),(f)). 
    }
    \label{fig:mix_results}
    % \vspace{-1em}
\end{figure}

The superior performance of UA-GAN can be illustrated by qualitative examples in Fig.~\ref{fig:mix_results}. On MNIST+Fashion dataset (subfigures a-c), the average aggregation strategy used by Avg-GAN could not effectively aggregate outputs of `non-identical' local discriminators. Therefore, it only learns to generate the common patterns, e.g., Fashion images (Fig.~\ref{fig:mix_results}(a)). MD-GAN fails to produce high quality images (Fig.~\ref{fig:mix_results}(b)), probably because the discriminator switch makes the training not stable enough. Meanwhile, our UA-GAN is able to generate the mixture with both common patterns (Fashion images) and site-specific images (different digits from MNIST) with high quality.
Similar phenomenon can be observed for MNIST+Font (subfigures d-f). Avg-GAN only learns the common pattern (computerized fonts from Font dataset), MD-GAN gives low quality images whereas UA-GAN can also learns the high-quality site-specific handwriting digits (MNIST). 
% The superior performance of UA-GAN can be illustrated by qualitative examples in Fig.~\ref{fig:mix_results}. On MNIST+Fashion dataset (subfigures a-c), the average aggregation strategy used by both Avg GAN and MD-GAN could not effectively aggregate outputs of `non-identical' local discriminators. Thus, these methods only learn to generate the common patterns, e.g., Fashion images. Meanwhile, UA-GAN is able to generate the mixture with both common patterns (Fashion images) and site-specific images (different digits from MNIST).
% Similar phenomenon can be observed for MNIST+Font (subfigures d-f). Average strategy only learns the common pattern (computerized fonts from Font dataset) whereas UA-GAN also learns the site-specific handwriting digits (MNIST). 

Note that we also compare the methods on mixture datasets with i.i.d local distributions, i.e., all local datasets are sampled in a same way from the real datasets. In an i.i.d setting, all federated GANs and the centralized GAN perform similarly. More results will be included in the supplemental material.

% \begin{figure}[t]
%     \centering
%      \begin{minipage}{0.16\linewidth}
%     \centering\includegraphics[width=\textwidth]{pics/fashion-identical-avg.png} \\ (a) Avg GAN
%     \end{minipage}
%     \begin{minipage}{0.16\linewidth}
%     \centering\includegraphics[width=\textwidth]{pics/fashion-identical-mdgan.png}  \\ (b) MD-GAN
%     \end{minipage}
%     \begin{minipage}{0.16\linewidth}
%     \centering\includegraphics[width=\textwidth]{pics/fashion-identical-ours.png} \\ (c)  Ours
%     \end{minipage}
%     \begin{minipage}{0.16\linewidth}
%     \centering\includegraphics[width=\textwidth]{pics/font-identical-avg.png}  \\ (d) Avg GAN
%     \end{minipage}
%     \begin{minipage}{0.16\linewidth}
%     \centering\includegraphics[width=\textwidth]{pics/font-identical-mdgan.png}  \\ (e) MD-GAN
%     \end{minipage}
%     \begin{minipage}{0.16\linewidth}
%     \centering\includegraphics[width=\textwidth]{pics/font-identical-ours.png} \\ (f) Ours
%     \end{minipage}
%     \caption{Synthetic images on the identical MNIST+fashionMNIST dataset ((a),(b),(c)) and MNIST+font dataset ((d),(e),(f)) using the average method, MD-GAN~\cite{hardy2019md} and our UA-GAN method. All of the models could capture distributions over all kinds of dataset, and generate diverse dataset.}
%     \label{fig:identical_results}
%     \vspace{-1em}
% \end{figure}
% \vspace{-.7em}
\section{Conclusion and Future Work}
% \vspace{-.7em}
In this work, we proposed a provably correct federated GAN. It simulates a centralized discriminator via carefully aggregating the feedbacks of all local discriminators. We proved that the generator learns the target distribution. We also analyzed the error bound when the discriminator is suboptimal due to local dataset limitation.
A well-trained federated GAN enpowers GANs to learn from diversified datasets. It can also be used as a data provider for training task-specific models without accessing or storing privacy sensitive data.

% \subsubsection*{Acknowledgments}
% Use unnumbered third level headings for the acknowledgments. All
% acknowledgments, including those to funding agencies, go at the end of the paper.

\bibliography{UA_GAN}
\bibliographystyle{plain}

\appendix
\setcounter{thm}{0}
\setcounter{lemma}{0}
\setcounter{obs}{0}
\setcounter{cor}{0}

In this supplemental material, we provide proofs of the theorems in the main paper (Sec.~\ref{sec:proof}). We also provide additional experimental details and results (Sec.~\ref{sec:exp}).	
\section{Proofs of Theorems in Section 3}
\label{sec:proof}
We recall the definition of odds value. 
	\begin{definition}[odds value]
	Given a probability $\phi \in (0,1)$, its odds value is $\Phi(\phi ) \triangleq \frac{\phi}{1-\phi}$. Note the definition requires $\phi\neq 1$. 
    \end{definition}

	\subsection{Analysis of Optimal Discriminator}
	We recall the theorem of the correctness of UA-GAN. 
		\begin{thm} [Correctness]
	    Suppose all discriminators $D_j$'s are optimal. $D_{ua}(x)$ is computed via Eq.~(\ref{grad_aggre_app}).  The optimal distribution of the Jenson-Shannon divergence loss:
	   % \cc{What loss? Jenson Shannon divergence? Should you explain more here? Or just assume ppl follows as it is folklore for GAN?}
	   % \vspace{-.1em}
	    \begin{equation}
	    \argmin\limits_{q}L(q)= \mathbb{E}_{x\sim p(x)} [\log D_{ua}(x)] +\mathbb{E}_{x\sim q(x)} [\log(1-D_{ua}(x)]
	    \end{equation}
        % \vspace{-.1em}
	    is  $q^*=p$ where $q$ is the density (mass) function of $G(z)$.
	\end{thm}
To prove the theorem, we first introduce the following Lemma, which is similar to Proposition 1 in~\cite{goodfellow2014generative}. We include the Lemma and Proof here for completeness.
		\begin{lemma}\label{lem1}
		When generator $G$ is fixed,  the optimal discriminator $D_j(x)$ is :\\
		\begin{equation}
		D_j(x)=\frac{p_j(x)}{p_j(x)+q(x)}
		\end{equation}
	\end{lemma}
	%	The proof could be found in \cite{goodfellow}, we include here for completeness.\\
	\textbf{Proof}:\\
	%%	 By taking derivative w.r.t $D_j(x)$ on $V(D,G)y$ we have $\sum \pi_j\int\limits_{y} s_j(x)\int\limits_{x} \frac{p_j(x)}{D_j(x)}  -q(x)log(1-D_j(x)) dydx$
	\begin{equation*} 
	\begin{aligned}
	&\max\limits_{D_j}V_j(D_j)=\max\limits_{D_j}\int\limits_{x} p_j(x)log D_j(x)+q(x)log(1-D_j(x)) dx\\
	&\leq \int\limits_{x} \max\limits_{D_j} \{p_j(x)log D_j(x)+q(x)log(1-D_j(x)) \}dx
	\end{aligned}
	\end{equation*}
	by setting $D_j(x)=\frac{p_j(x)}{p_j(x)+q(x)}$ we can maximize each component in the integral thus make the inequality hold with equality. \qed

% 	\begin{thm}
% 	    Suppose all discriminators $D_j(x), j\in [K]$ always behave optimally and $D_{ua}(x)$ is computed via equation (\ref{grad_aggre_app}). The optimal distribution of loss:\\
% 	    $$\min\limits_{G}V(G)= \mathbb{E}_{x\sim p(x)} [\log D_{ua}(x)] +\mathbb{E}_{z\sim \mathcal{N}(0,1)} [\log(1-D^*(G(z)|y)]$$

% 	    is  $q(G(z))=p(x)$ where $q(G(z))$ is density (mass) function of $G(z)$.
% 	\end{thm}
% 	The theorem shares a similar spirit with Theorem 1 in 

	\textbf{Proof of Theorem \ref{Unbiased}}:
	Suppose in each training step the discriminator achieves its maximal criterion in Lemma \ref{lem1}, the simulated $D_{ua}(x)$ becomes:\\
		\begin{equation} \label{grad_aggre_app}
	D_{ua}(x)=\frac{\sum_{j}^K  \pi_j (y) \frac{D^*_j(x)}{1-D^*_j(x)}}{1+\sum_{j}^K \pi_j (y) \frac{D^*_j(x)}{1-D^*_j(x)}}
	\end{equation} 
		Given that discriminators $D_1,...,D_K$ behave optimally, the value of $\frac{D_j(x)}{1-D_j(x)}=\frac{p_j(x)}{q(x)}$ which implies $\sum_{j} \frac{ \pi_j D_j(x)}{1-D_j(x)}=\frac{\sum_{j}  \pi_j  p_j(x)}{q(x)}=\frac{D_{ua}(x)}{1-D_{ua}(x)}$. By the aggregation formula \ref{grad_aggre_app}, the simulated discriminator will be $D_{ua}(x)=\frac{\sum_{j} \pi_j p_j(x)}{\sum_{j} \pi_j p_j(x)+q(x)}=\frac{p(x)}{p(x)+q(x)}$. 
		Suppose in each training step the discriminator achieves its maximal criterion in Lemma \ref{lem1}, the loss function for the generator becomes:
	\begin{equation*}
	\begin{aligned}
	&\min\limits_{a}L(q)= \mathbb{E}_{x\sim p(x)} [\log D(x)] +\mathbb{E}_{\hat{x}\sim q(\hat{x}|y)} [\log(1-D(\hat{x})]\\
	&= \mathbb{E}_{x\sim p(x} [\log D(x)]+\mathbb{E}_{\hat{x}\sim q(\hat{x})} [\log (1-D(\hat{x}))]\\
	&= \int\limits_{x} p(x)\log \frac{p(x)}{p(x)+q(x)}+q(x)\log \frac{q(x)}{p(x)+q(x)} dx
	\end{aligned}
	\end{equation*}
	The above loss function has optimal solution of $q$ due to Theorem 1 in \cite{goodfellow2014generative}.  \qed
	
% 	\begin{remark}
%     	The only information needs to be transmitted to generator is the score $D(\hat{x})$ and corresponding gradient on each synthetic image which  is computed by aggregating information from different discriminators according to Equation \ref{aggre}. Note that the score on real images $D(x)$ does not need to be computed.
%     \end{remark}

% 	\begin{remark}
% 	Note that the only information needs to be transmitted to generator is the score $D(\hat{x})$ and corresponding gradient on each synthetic image which could be computed by aggregating information from different discriminators. Note that the score on real images $D(x)$ does not need to be computed as long as every player $\{D_1,...,D_K\}$ plays optimally thus there is no transmission between $\{D_1,...,D_K\}$. 
% 	\end{remark}
	\subsection{Analysis of sub-optimal discriminator}
	We provide proofs of Theorems \ref{main}, \ref{tightness}, and Corollary \ref{main_cor}. 
		\begin{thm}[Suboptimality Bound for a Single Discriminator] 
		Suppose a discriminator $\widetilde{D}(x)$ is a perturbed version of the optimal discriminator $D(x)$, s.t.~$\Phi(\widetilde{D}(x))= \Phi(D(x)) \xi(x)$ with  $|1-\xi(x)| \leq \delta$ and $\delta \leq 1/8$. 
		Let $q^\ast$ be the optimal distribution of the Jensen-Shannon divergence loss based on the perturbed discriminator 
		\begin{equation}
	    q^\ast = \argmin\limits_{q}L(q)= \mathbb{E}_{x\sim p(x)} [\log \widetilde{D}(x)] +\mathbb{E}_{x\sim q(x)} [\log(1-\widetilde{D}(x)].
	    \end{equation}
	    Then $q^\ast$ satisfies
	 $|q^\ast(x)/p(x)-1| \leq 16 \delta, \forall x$.
	\end{thm}

	\begin{lemma}\label{log_lem}
		Suppose $0<|a|,|b| \leq \frac{1}{8}$, $\log \rho =\log(\frac{1}{2}+a)+b$ with $0<\rho<1$, we have $1-2|a|-2|b| \leq \frac{\rho}{1-\rho} \leq 1+4|a|+4|b|$. 
	\end{lemma}
	
	\textbf{Proof}:\\
	In the proof we will use following fact:
	\begin{equation}\label{fact_1}
		1+x \leq e^x \leq 1+2x, \text{ for } 0<x<\frac{1}{8}
	\end{equation}
	By equation $\log \rho =\log(\frac{1}{2}+a)+b$ we have $\rho =(\frac{1}{2}+a)e^b$ thus:
	\begin{equation}
		\begin{aligned}
			&(\frac{1}{2}-|a|)(1-2|b|) \leq \rho \leq (\frac{1}{2}+|a|)(1+2|b|)\\
			 &\frac{1}{2}-|a|-|b| \leq \rho \leq 	\frac{1}{2}+|a|+|b|+2|ab|\\
		    &\frac{1}{2}-|a|-|b|-2|ab|  \leq 1-\rho \leq 	\frac{1}{2}+|a|+|b| \\
			 &\frac{\frac{1}{2}-|a|-|b| }{\frac{1}{2}+|a|+|b|}  \leq \frac{\rho}{1-\rho} \leq 
			 \frac{\frac{1}{2}+|a|+|b|+2|ab|}{\frac{1}{2}-|a|-|b|-2|ab|}\\
			 &(1-2|a|-2|b|)  \leq \frac{\rho}{1-\rho} \leq  (1+4|a|+4|b|)
		\end{aligned}
	\end{equation}
	\qed
	
	\begin{lemma}\label{convexity}
		Suppose $h(x)/p(x) \geq \frac{1}{2} $, the following  loss function is strongly convex:\\
		\begin{equation}
			L(q)=\int_{x} p(x) \log\frac{h(x)}{h(x)+q(x)}+q(x)\log\frac{q(x)}{q(x)+h(x)} dx
		\end{equation}
	\end{lemma}

	\textbf{Proof}:\\
	The first order derivative of $L(q)$ is $\frac{\partial L(q)}{\partial q(x)}=\frac{h(x)-p(x)}{q(x)+h(x)} + \log\frac{q(x)}{q(x)+h(x)}$. The second order derivative is $\frac{\partial^2 L(q)}{\partial q(x)^2}= \frac{ q(x)(2h(x)-p(x))+h(x)^2}{(q(x)+h(x))^2q(x)}$. (There is no non-diagonal elements in the Hessian). \qed
	
	The proof for Theorem \ref{main} is shown below. The theorem focuses on Jensen-Shannon Divergence loss \cite{guha2007sketching,lin1991divergence}.  We stress that the analysis is general and works for both general GAN and conditional GAN.   
% 	\begin{thm} \label{main}
% 		Suppose a discriminator $\tilde{D}(x)$ is a perturbed version of discriminator $D(x)$ s.t. $\frac{D(x)}{1-D(x)}=\frac{p(x)}{q(x)}$ is perturbed by $\xi(x)$ with  $1-\delta \leq  \xi(x) \leq 1+\delta$ with $\delta \leq \frac{1}{8}$, then optimal distribution satisfies $|\frac{q(x)}{p(x)}-1| \leq 16 \delta, \forall x$.
% 	\end{thm}

	\textbf{Proof of Theorem \ref{main}}:\\
	Note $\Phi(D(x))=\frac{p(x)}{q(x)}$. The $\xi(x)$ perturbed odds value gives $\Phi(\widetilde{D}(x))= \Phi(D(x)) \xi(x)=\frac{p(x)\xi(x)}{q(x)}$. Let $h(x)=p(x)\xi(x)$, we have $1-\delta \leq \frac{h(x)}{p(x)} \leq 1+\delta$ and 
	$1-\delta \leq \frac{p(x)}{h(x)} \leq 1+2\delta$. The perturbed value of discriminator will be $\widetilde{D}(x)=\frac{h(x)}{h(x)+q(x)}$. The loss function that the generator distribution $q(x)$ seeks to minimize  a perturbed Jensen-Shannon Divergence loss:\\
	\begin{equation}
		L(q)=\int_{x} p(x)\log \frac{h(x)}{h(x)+q(x)}+q(x)\log\frac{q(x)}{h(x)+q(x)} dx + \lambda (\int_{x} q(x) dx -1).
	\end{equation}
	where $\lambda$ represents the Lagrangian Multiplier.  
	In Lemma \ref{convexity} we verify the convexity of this loss function with regulated behavior of $h(x)$. The derivative of $L(q)$ w.r.t. $q(x)$ is:
	\begin{equation*}
		-\frac{p(x)}{q(x)+h(x)} + \log(q(x))+1-\log(q(x)+h(x)) -\frac{q(x)}{q(x)+h(x)}+\lambda
	\end{equation*}
	which needs to be $0$ for all value of $x$ \cite{boyd2004convex}. Thus we have following equation:\\
	\begin{equation} \label{grad_eq}
	\frac{h(x)-p(x)}{q(x)+h(x)} + \log\frac{q(x)}{q(x)+h(x)}=-\lambda
	\end{equation}
	Since above equation holds for all $x$. Let $\frac{h(x)-p(x)}{q(x)+h(x)}=\Delta(x)$ which has value bounded by $-\delta \leq \Delta(x) \leq \delta$. we can multiply $p(x)+q(x)$ on both side and integral over $x$:\\
	\begin{equation*}
	\int_{x} (h(x)+q(x))\Delta(x) + (h(x)+q(x))\log\frac{q(x)}{q(x)+h(x)} dx=-\int_x \lambda(h(x)+q(x)) dx
	\end{equation*}
	which gives us $\Delta_1+\Delta_2 =-2 \lambda$ where:\\
	\begin{equation*}
	 \Delta_1=	\frac{1}{2}\int_{x} (h(x)+q(x))\Delta(x) dx , \;\;\Delta_2= \frac{1}{2}\int_{x} (h(x)+q(x))\log\frac{q(x)}{q(x)+h(x)} dx
	\end{equation*}
	Plugging in above derived value of $\lambda$ back into Equation \ref{grad_eq} we have:\\
	\begin{equation} \label{log_eq}
		\log\frac{q(x)}{q(x)+p(x)}=\Delta_1+\Delta_2-\Delta(x)
	\end{equation}
	By the uniform upper bound on  $|\Delta(x)| \leq 2\delta$ we have $ 1-\frac{3}{2}\delta \leq e^{\Delta_1} \leq 1+\frac{3}{2}\delta$ and $ 1-\frac{3}{2}\delta \leq e^{\Delta} \leq 1+\frac{3}{2}\delta$. By taking exponential operation on both sides of Equation \ref{log_eq} we have :\\
	\begin{equation*}
		\frac{q(x)}{q(x)+h(x)}= e^{\Delta_2}*(1\pm 4\delta)
	\end{equation*}
	Thus we can bound the value of $e^{\Delta_2}$ by the equation:\\
	\begin{equation} \label{eq_exp_1}
		q(x)=(q(x)+h(x))(e^{\Delta_2}*(1\pm 4\delta))
	\end{equation}
	 . Due to the fact that the equation holds for all value of $x$, by integrating Equation \ref{eq_exp_1} we have $e^{\Delta_2} =\frac{1}{2} (1\pm 4\delta )$ which is equivalent to  $\Delta_2 =\log(\frac{1}{2}\pm 4\delta)$. Thus $\log\frac{q(x)}{q(x)+h(x)} = \log(\frac{1}{2}\pm 4\delta)\pm \Delta$. By Lemma \ref{log_lem}, we have $\frac{q(x)}{p(x)} = 1\pm 16\delta$. \qed

% 	 \begin{thm}[Tightness of bound in Theorem 3] \label{tightness}
% 		Suppose a discriminator $\tilde{D}(x)$ is a perturbed version of discriminator $D(x)$ s.t. $\frac{D(x)}{1-D(x)}=\frac{p(x)}{q(x)}$ is perturbed by $\xi(x)$ with  $|\xi(x)-1|= \delta(x)$ with $\gamma \leq \delta(x) \leq \frac{1}{8}$, then optimal distribution satisfies $|\frac{q(x)}{p(x)}-1| \geq \frac{\gamma}{16}, \forall x$.
%     \end{thm}
    	  \begin{thm}[Tightness of the Bound in Theorem (\ref{main})]
	  Given a perturbed discriminator $\widetilde{D}(x)$ of the optimal one $D(x)$, s.t.~$\Phi(\widetilde{D}(x))= \Phi(D(x))\xi(x)$ 
% 		Suppose a discriminator $\tilde{D}(x)$ is a perturbed version of discriminator $D(x)$ s.t. $\frac{D(x)}{1-D(x)}=\frac{p(x)}{q(x)}$ is perturbed by $\xi(x)$ 
		with  $|\xi(x)-1| \geq \gamma$ and $\gamma \leq 1/8$. The optimal distribution $q^\ast$ as in Eq.~(\ref{eq:jenson-shannon-2}) satisfies $|q^\ast(x)/p(x)-1| \geq \gamma/16$, $\forall x$.
    \end{thm} 
    
    \textbf{Proof}:\\
        By similar steps to proof in Theorem 3 we have:
    	\begin{equation}
    		L(q)=\int_{x} p(x)\log \frac{h(x)}{h(x)+q(x)}+q(x)\log\frac{q(x)}{h(x)+q(x)} dx + \lambda (\int_{x} q(x) dx -1).
    	\end{equation}
    	By setting the derivative of $L(q)$ w.r.t. $q(x)$ be $0$ we have :
    	\begin{equation}
    	\frac{h(x)-p(x)}{q(x)+h(x)} + \log\frac{q(x)}{q(x)+h(x)}=-\lambda
    	\end{equation}
    	Let $\frac{h(x)-p(x)}{q(x)+h(x)}=\Delta(x)$. By assumption that $|\xi(x)-1| \geq \gamma$ we have  $$|\Delta(x)|=\frac{|h(x)-p(x)|}{q(x)+h(x)}=\frac{|\xi(x)-1|}{\frac{h(x)}{p(x)}+\frac{q(x)}{p(x)}} $$.
    	Due to the fact that $|\xi(x)-1|\leq \frac{1}{8}$, we have $\frac{q(x)}{p(x)} \leq 2$. Thus we can bound $|\Delta(x)| \geq \frac{\gamma}{8}$.
    % 	\begin{equation*}
    % 	\begin{aligned}
    % 	    &|\Delta(x)|=\frac{|h(x)-p(x)|}{q(x)+h(x)} \\
    % 	    &
    % 	\end{aligned}
    % 	\end{equatio*}

    	Next we analyze following equation, we can multiply $p(x)+q(x)$ on  both side of (\ref{grad_eq}) and integral over $x$:\\
    	\begin{equation*}
    	\int_{x} (h(x)+q(x))\Delta(x) + (h(x)+q(x))\log\frac{q(x)}{q(x)+h(x)} dx=-\int_x \lambda(h(x)+q(x)) dx
    	\end{equation*}
    	which gives us $\Delta_1+\Delta_2 =-2 \lambda$ where:\\
    	\begin{equation*}
    	 \Delta_1=	\frac{1}{2}\int_{x} (h(x)+q(x))\Delta(x) dx , \;\;\Delta_2= \frac{1}{2}\int_{x} (h(x)+q(x))\log\frac{q(x)}{q(x)+h(x)} dx
    	\end{equation*}
    	Plugging in above derived value of $\lambda$ back into Equation \ref{grad_eq} we have:\\
    	\begin{equation}
    		\log\frac{q(x)}{q(x)+p(x)}=\Delta_1+\Delta_2-\Delta(x)
    	\end{equation}
        Next we analyze the term $\Delta_1-\Delta(x)$. 
        \begin{equation*}
        \begin{aligned}
            \Delta_1-\Delta(x)=	&\frac{1}{2}\int_{x} (h(x)+q(x))\Delta(x) dx -\Delta(x)\\
            &=\frac{1}{2}\int_{x} (h(x)+q(x))\frac{h(x)-p(x)}{q(x)+h(x)} dx-\Delta(x)\\
            &=\Delta(x)
        \end{aligned}
        \end{equation*}
        Due to the fact that $|\Delta(x)|> \frac{\gamma}{8}$, We can derive that $|\frac{e^{\Delta_2}}{\frac{1}{2}} -1|> \frac{\gamma}{16}$.
        By an analysis similar to Theorem 3 we have $|\frac{q(x)}{p(x)}-1|\geq \frac{\gamma}{64}$.  \qed

	 \begin{cor} [Suboptimality Bound for UA-GAN]
	 	Assume suboptimal local discriminators $\widetilde{D_j}(x)$ are the perturbed versions of the optimal ones $D_j(x)$. And the suboptimality is bounded as: $\Phi(\widetilde{D_j}(x))=\Phi(D_j(x))\xi_j(x)$  with  $|\xi_j(x)-1| \leq \delta \leq 1/8$, $\forall x$. The centralized discriminator $\widetilde{D_{ua}}(x)$ is computed using these perturbed local discriminators such that $\Phi(\widetilde{D_{ua}}(x))=\sum_{j=1}^K\pi_j\Phi(\widetilde{D_j}(x))$. 
% 	 	Let $q(x)$ be the density (mass) function of $G(z)$. 
		Let $q^\ast$ be the optimal distribution of the Jensen-Shannon divergence loss based on the perturbed UA discriminator $\widetilde{D_{ua}}$ 
		\begin{equation}
	    q^\ast = \argmin\limits_{q}L(q)= \mathbb{E}_{x\sim p(x)} [\log \widetilde{D_{ua}}(x)] +\mathbb{E}_{x\sim q(x)} [\log(1-\widetilde{D_{ua}}(x)].
	    \end{equation}
	    Then $q^\ast$ satisfies $|q^\ast(x)/p(x)-1|=O(\delta)$.   In particular, the optimal distribution $q^*(x)$ has $O(\delta)$ total variation distance to the target distribution $p(x)$.
	 	
	 \end{cor}  
	 \textbf{Proof}:\\
	 Let $v_j$'s be odds values of optimal discriminators $D_j(x)$'s: $v_j=\frac{D_j(x)}{1-D_j(x)}$ and $\widetilde{v_j}$'s be odds values of suboptimal discriminators $\widetilde{D_j}(x)$'s. It suffices to show
	 $|\frac{\sum_{j}\pi_jv_j}{\sum_{j}\pi_j\tilde{v_j}}-1| \leq \delta$ and apply Theorem \ref{main}. \qed

    \section{Additional Experimental Details and Results} \label{sec:exp}

\begin{algorithm}[!ht]
\caption{Precise Training Algorithm of UA-GAN.}
 \begin{algorithmic}[1]\label{alg:UA-GAN_pre}
%  \label{alg:AGF}
\STATE {\bfseries Input:} Batch size $m$, datasets $\{\mathcal{D}_j\}$, size of datasets $\{\pi_j=\frac{n_j}{n}\}$. 
\STATE {{\bfseries Output:} $G$, $D_j, \forall j \in [K]$.}
  \FOR{Number of total training iterations }
%   \STATE{-- Sample $y_i, i \in [m]$ according to  $supp(y)$ and  $\pi_j(y), j\in[K]$.}
\FOR{Number of iterations to train discriminator }
\STATE \COMMENT{Work at the central server.}
    % \FOR{$=1,\cdots,T$}
  \STATE{$G$ generates synthetic data: $\hat{x}_i = G(z_i)$, $i=1,\cdots,m$. }
  \STATE{Send batch of synthetic data $\mathcal{D}_{syn}=\{\hat{x}_1,\cdots,\hat{x}_m\}$ to all $K$ sites.}
  \FOR{$j = 1, \cdots, K$}
  \STATE \COMMENT{Work at each local site.}
  \STATE{Uniformly randomly choose $m$ real samples $ \{x^j_1,\cdots,x^j_m\}$ from $\mathcal{D}_j$: }
  \STATE{Update the parameters of local discriminator $D_j$: $\theta_j$ using $$\nabla_{\theta_j} \frac{1}{m} \sum_{i=1}^{m} \left[ \log (D_j(x^j_i))+ \log(1-D_j (\hat{x_i}))) \right]$$}
%   \STATE{Update the local discriminator, $D_j$, using real samples from $ \mathcal{D}_j$ and synthetic data batch, $\mathcal{D}_{syn}$, based on Eq.~\ref{Dis}.}
    \ENDFOR

\ENDFOR
   \STATE \COMMENT{Work at each local site.}
    \STATE{$G$ generates synthetic data: $\hat{x}_i = G(z_i)$, $i=1,\cdots,m$. }
  \STATE{Send batch of synthetic data $\mathcal{D}_{syn}=\{\hat{x}_1,\cdots,\hat{x}_m\}$ to all $K$ sites.}
     \FOR{$j = 1, \cdots, K$}
    \STATE \COMMENT{Work at each local site.}
    \STATE{Output predictions and gradients for synthetic data $D_j(\hat{x}_i)$, $\partial D_j(\hat{x}_i) / \partial \hat{x}_i$, $i =1, \cdots,m$. Send them to the central server.}  
    \ENDFOR
  \STATE \COMMENT{Work at the central server.}
  \STATE{Simulate value of $D_{ua}(\hat{x}_i)$ via Eq.~(\ref{aggre}), $\forall i$.}
%   \STATE{Update $G$ based on Eq.~\eqref{Gen}, using gradients from $D_j$'s.}
  \STATE{Update parameter of $G$: $\theta_G$  by descending its stochastic gradient:
  $$ \frac{1}{m} \sum_{i=1}^m \frac{\partial \log(1-D_{ua}(\hat{x}_i))}{\partial D_{ua}(\hat{x}_i)}\frac{\partial D_{ua}}{\partial \Phi(D_{ua}) } \sum_{j=1}^{K}\left[ \frac{\partial \Phi(D_{ua})}{\partial D_j(\hat{x}_i)} \frac{\partial D_j(\hat{x}_i)}{\partial \hat{x}_i}\right]\frac{\partial \hat{x}_i}{\partial\theta_G}$$  }
  \ENDFOR
  \\The gradient-based updates can use any standard gradient-based learning rule.
 \end{algorithmic}
\end{algorithm}

\begin{figure}[!ht]
    \centering
     \begin{minipage}{0.45\linewidth}
    \centering\includegraphics[width=\textwidth]{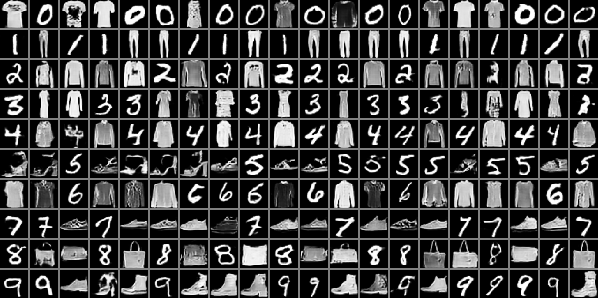} \\ (a) Avg GAN MNIST+Fashion
    \end{minipage}
    \begin{minipage}{0.45\linewidth}
    \centering\includegraphics[width=\textwidth]{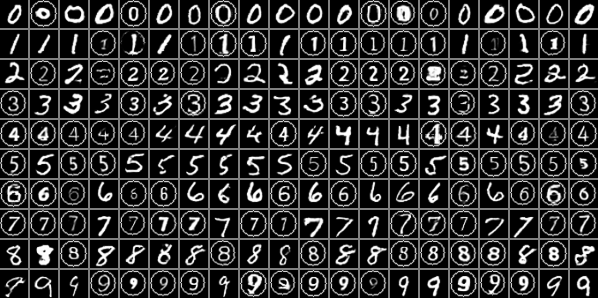}  \\ (b) Avg GAN MNIST+Font
    \end{minipage}
    \begin{minipage}{0.45\linewidth}
    \centering\includegraphics[width=\textwidth]{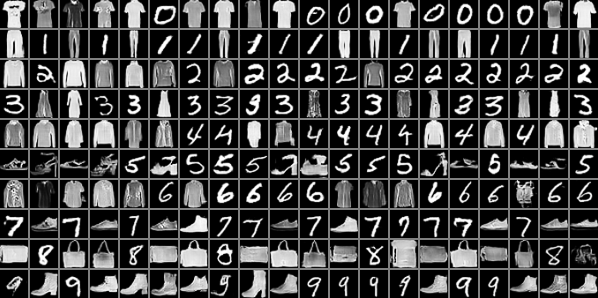} \\ (c)  MD-GAN MNIST+Fashion
    \end{minipage}
    \begin{minipage}{0.45\linewidth}
    \centering\includegraphics[width=\textwidth]{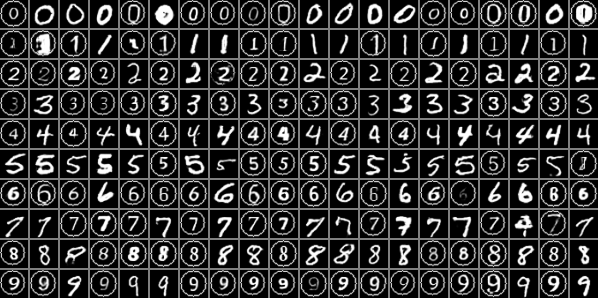}  \\ (d) MD-GAN MNIST+Font
    \end{minipage}
    \begin{minipage}{0.45\linewidth}
    \centering\includegraphics[width=\textwidth]{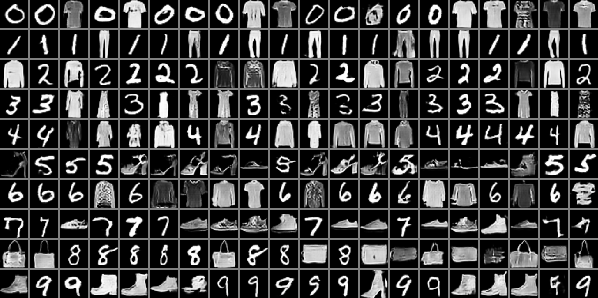}  \\ (e) UA-GAN MNIST+Fashion
    \end{minipage}
    \begin{minipage}{0.45\linewidth}
    \centering\includegraphics[width=\textwidth]{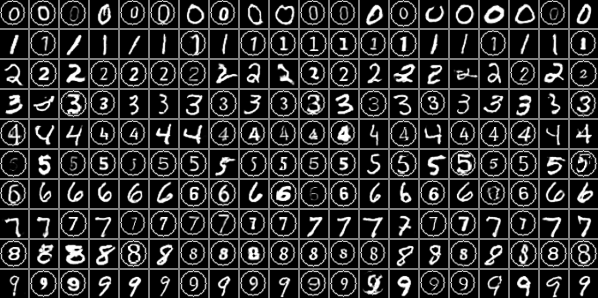} \\ (f) UA-GAN MNIST+Font
    \end{minipage}
    \caption{Synthetic images on the identical MNIST+Fashion dataset ((a),(c),(e)) and MNIST+Font dataset ((b),(d),(f)) using the average method, MD-GAN~\cite{hardy2019md} and our UA-GAN method. All of the models could capture distributions over  MNIST and Fashion/Font. }
    \label{fig:identical_results}
    % \vspace{-1em}
\end{figure}

% \subsection{}
% \myparagraph{Implementation details}

\paragraph{Implementation Details:}Here we summarize details of the network we use in the experiments. Our UA-GAN has one centralized generator and multiple local discriminators.
The generator consists of two fully-connected layers (for input noise and label, respectively), five residual blocks~\cite{he2016resnet} and three upsampling layers. Each discriminator has two convolutional layers (for image and label, respectively), five residual blocks and three average pooling layers. LeakyReLU activation is used in both generator and discriminators. During training, we apply $1$ gradient update of  the discriminators in each round. Each model is trained with Adam optimizer for 400 epochs with a batch size of 256. The learning rate is initially 0.0002 and linear decays to 0 from epoch 200.
The VGG \cite{simonyan2014very} 11-layer model is used for the downstream classification task. We pad the image to $32 \times 32$ and then randomly crop them to $28\times28$ with a batch size of 64 as input. The model is trained with SGD optimizer using a learning rate of 0.01 for 150 epochs.

% \subsection{}
% \vspace{-.5em}
\paragraph{Dataset Details:}One of our foundational datasets is the Font dataset. It is created from 2500+ fonts of digits taken from the Google Fonts database. 
Similar to MNIST, it consists of 10 classes of $28\times28$ grayscale images, with 60k samples for training and 29k samples for test.

Based on the MNIST, Fashion-MNIST and Font dataset, we create both i.i.d mixture datasets and non-identical datasets. Details on non-identical datasets have been provided in the main paper. Here we provide details on two i.i.d datasets.
 (1) i.i.d MNIST+Fashion; (2) i.i.d MNIST+Font. Each of the 10 distributed sites  contains 10\% of mixture dataset which is uniformly sampled (without replacement) from MNIST and Fashion/Font. 
 
%  \vspace{-.3em}
\subsection{Empirical Results on I.I.D.~Datasets}
The quantitative results on the i.i.d mixture datasets are shown in Table \ref{tab:acc_iid}. One can see all three distributed GAN methods have comparable performance. It can also be observed from qualitative examples in Fig.~\ref{fig:identical_results} that all three methods achieve similar results. This suggests that all three approaches can be used to train distributed GAN when datasets have i.i.d.~distribution e.g., the data is uniformly shuffled before sent to each discriminator. Note that with a similar performance, the UA-GAN has much smaller communication cost compared to MD-GAN since the UA-GAN does not  swap model parameters during training process. 

\begin{table}[!ht]
	\begin{center}
% 	\vspace{-.1em}
		\begin{tabular}{lcccccc}
		    \midrule
		    \multirow{2}{*}{Dataset} 
		    & \multicolumn{3}{c}{i.i.d Mnist + Fashion}
		    & \multicolumn{3}{c}{i.i.d Mnist + Font} \\ 
		    \cmidrule{2-7}
		    & Accuracy$\uparrow$ & IS$\uparrow$ & FID$\downarrow$ & Accuracy$\uparrow$ & IS$\uparrow$ & FID$\downarrow$ \\ 
		    \midrule
		    
			Real & 0.943 & 3.620 $\pm$ 0.021 & 0
			& 0.994 & 2.323 $\pm$ 0.011 & 0 \\ 
			\midrule
			Centralized GAN & 0.904  & 3.437 $\pm$ 0.021 & 8.35 & 0.979 & 1.978 $\pm$ 0.009 & 17.62 \\
			
			Avg GAN & 0.905 & 3.371 $\pm$ 0.026 & 12.83 & 0.967 & 1.923 $\pm$ 0.006  & 19.31 \\
			
			MD-GAN & 0.884 & 3.364 $\pm$ 0.024 & 13.63 & 0.971  & 1.938 $\pm$ 0.008 & 19.65   \\
			\midrule
			\textbf{UA-GAN} & 0.908 & 3.462 $\pm$ 0.024
 & 11.82 & 0.970 & 1.934 $\pm$ 0.008 & 19.18\\
			\bottomrule
		\end{tabular}
% 			\vspace{-.05in}
	\end{center}
	\caption{The classification accuracy and IS, FID scores on two i.i.d mixture datasets. All of the three architecture could learn the right distribution with i.i.d datasets. }	
%	\vspace{-0.1in}
	\label{tab:acc_iid}
% 	\vspace{-.2em}
\end{table}

% \vspace{-.3em}
\subsection{Additional Empirical Results on Non-identical Distribution}
\begin{figure}[ht]
    \centering
     \begin{minipage}{0.45\linewidth}
    \centering\includegraphics[width=\textwidth]{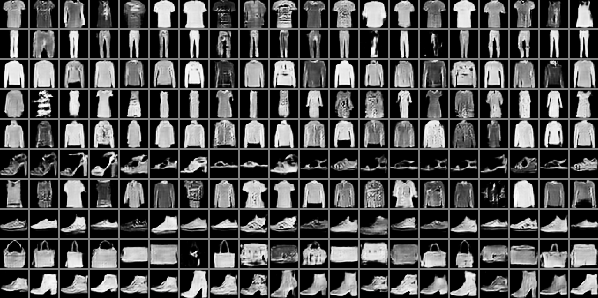} \\ (a) Avg GAN MNIST+Fashion
    \end{minipage}
    \begin{minipage}{0.45\linewidth}
    \centering\includegraphics[width=\textwidth]{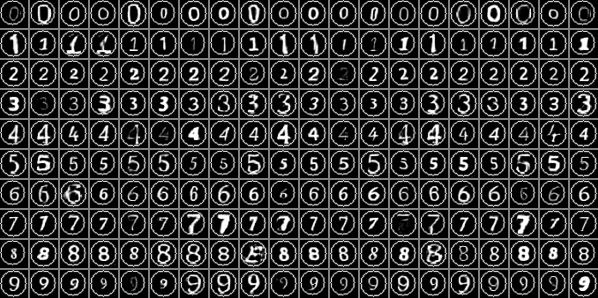}  \\ (b) Avg GAN MNIST+Font
    \end{minipage}
    \begin{minipage}{0.45\linewidth}
    \centering\includegraphics[width=\textwidth]{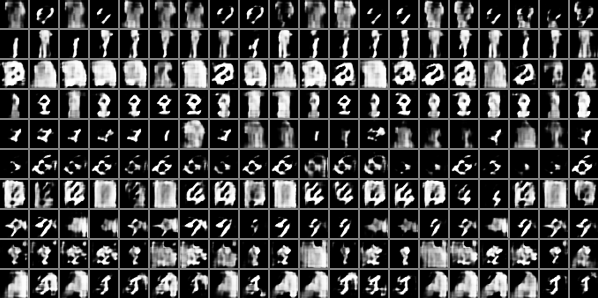} \\ (c)  MD-GAN MNIST+Fashion
    \end{minipage}
    \begin{minipage}{0.45\linewidth}
    \centering\includegraphics[width=\textwidth]{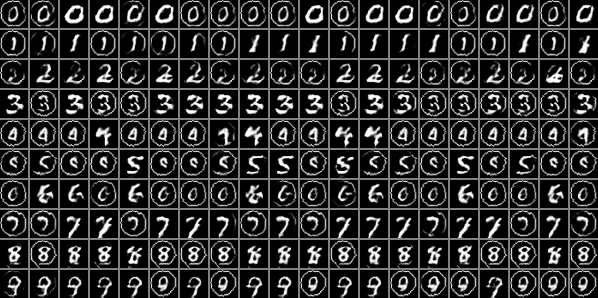}  \\ (d) MD-GAN MNIST+Font
    \end{minipage}
    \begin{minipage}{0.45\linewidth}
    \centering\includegraphics[width=\textwidth]{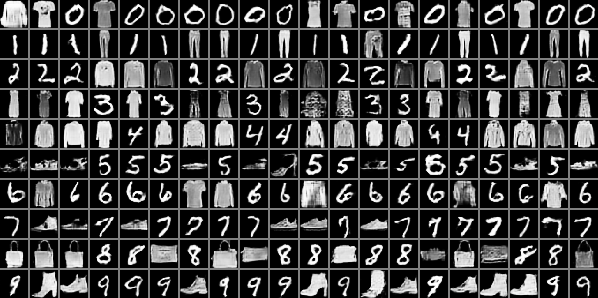}  \\ (e) UA-GAN MNIST+Fashion
    \end{minipage}
    \begin{minipage}{0.45\linewidth}
    \centering\includegraphics[width=\textwidth]{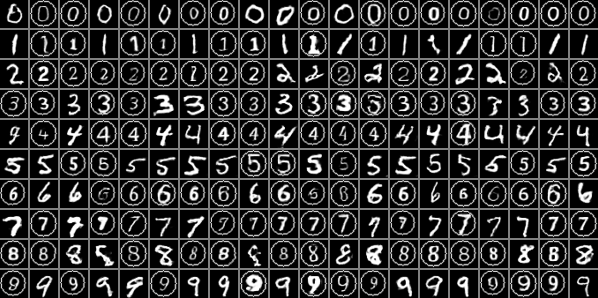} \\ (f) UA-GAN  MNIST+Font
    \end{minipage}
    \caption{Additional synthetic images on the non-identical MNIST+Fashion dataset ((a),(c),(e)) and MNIST+Font dataset ((b),(d),(f)) using the average method, MD-GAN~\cite{hardy2019md} and our UA-GAN method.  }
    \label{fig:non_identical_results}
    % \vspace{-1em}
\end{figure}

We provide additional 
synthetic images in non-identical distribution cases. See Fig.~\ref{fig:non_identical_results}. By using average aggregation method, the synthetic image produced by Avg GAN and MD GAN only have Fashion images in (a), (c)  and  Font images in (b), (d). Our method in (e) and (f) could capture different patterns in MNIST + Fashion/Font and generate diverse images.

\subsection{Empirical Results of Mixing Three Datasets}
\begin{figure}[ht]
    \centering
     \begin{minipage}{0.45\linewidth}
    \centering\includegraphics[width=\textwidth]{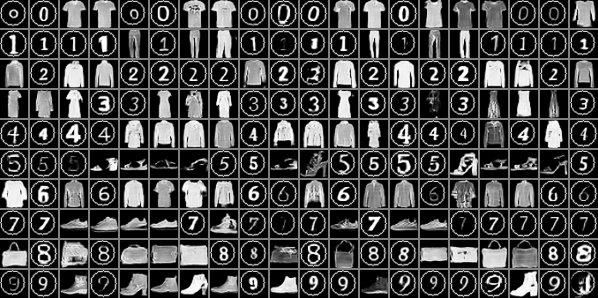} \\ (a) Avg-GAN MNIST+Fashion+Font
    \end{minipage}
    \begin{minipage}{0.45\linewidth}
    \centering\includegraphics[width=\textwidth]{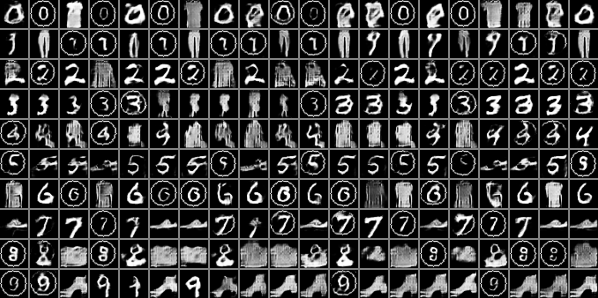}  \\ (b) MD-GAN MNIST+Fashion+Font
    \end{minipage}
    \begin{minipage}{0.45\linewidth}
    \centering\includegraphics[width=\textwidth]{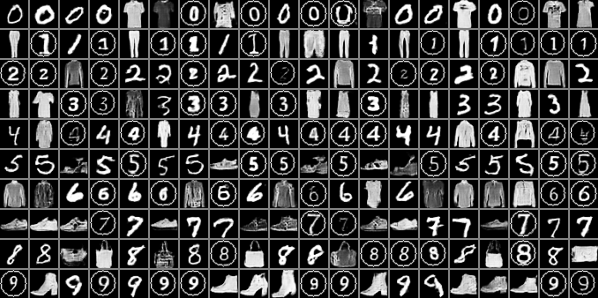} \\ (c)  UA-GAN MNIST+Fashion+Font
    \end{minipage}
    \caption{Synthetic images on the non-identical MNIST+Fashion+Font dataset using the average method, MD-GAN~\cite{hardy2019md} and our UA-GAN method.  }
    \label{fig:noniid-threedatasets}
\end{figure}

\begin{figure}[ht]
    \centering
     \begin{minipage}{0.45\linewidth}
    \centering\includegraphics[width=\textwidth]{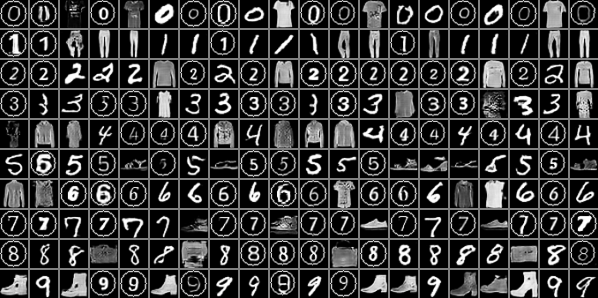} \\ (a) Avg-GAN MNIST+Fashion+Font
    \end{minipage}
    \begin{minipage}{0.45\linewidth}
    \centering\includegraphics[width=\textwidth]{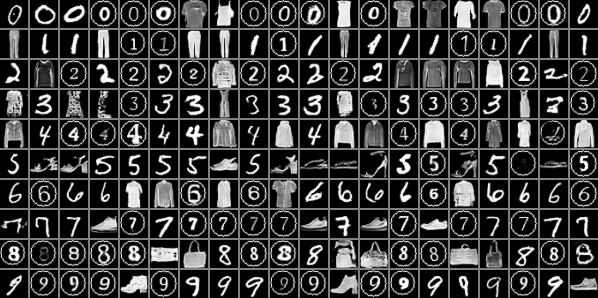}  \\ (b) MD-GAN MNIST+Fashion+Font
    \end{minipage}
    \begin{minipage}{0.45\linewidth}
    \centering\includegraphics[width=\textwidth]{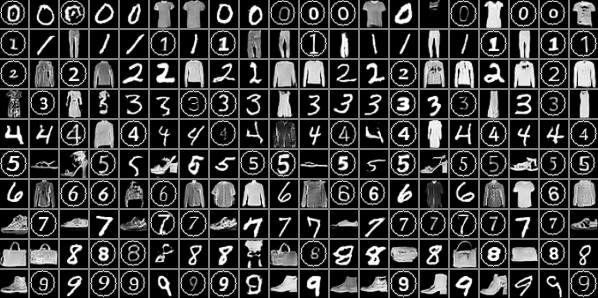} \\ (c)  UA-GAN MNIST+Fashion+Font
    \end{minipage}
    \caption{Synthetic images on the identical MNIST+Fashion+Font dataset using the average method, MD-GAN~\cite{hardy2019md} and our UA-GAN method.  }
    \label{fig:iid-threedatasets}
\end{figure}

We report the results of mixing the three datasets MNIST, FashionMNIST and Font. In the non-identical setting, we add MNIST data with a distinct class among 0$\sim$9. These data are distinguishable features for different sites. And we uniformly sample Fashion and Font data for all 10 distributed sites. These are common patterns across all sites. In the identical setting, all three datasets are uniformly distributed across the 10 sites. The quantitative results are shown in Table~\ref{tab:threedatasets}. The synthtic images are shown in Fig.~\ref{fig:noniid-threedatasets} and Fig.~\ref{fig:iid-threedatasets}. By using average aggregation method, the synthetic image produced by Avg-GAN and MD -GAN only have Fashion and Font images in Fig.~\ref{fig:noniid-threedatasets}(a), (b) . Our method in Fig.~\ref{fig:noniid-threedatasets}(c) could capture different patterns in MNIST + Fashion + Font and generate diverse images.

\begin{table}[!ht]
	\begin{center}
% 	\vspace{-.1em}
		\begin{tabular}{l|ccc|ccc}
		    \midrule
		    \multirow{2}{*}{Dataset} 
		    & \multicolumn{3}{c|}{non-i.i.d}
		    & \multicolumn{3}{c}{i.i.d} \\ 
		    \cmidrule{2-7}
		    & Accuracy$\uparrow$ & IS$\uparrow$ & FID$\downarrow$ & Accuracy$\uparrow$ & IS$\uparrow$ & FID$\downarrow$ \\ 
		    \midrule
			Real & 0.955 & 3.426 $\pm$ 0.023 & 0 & 0.955 & 3.426 $\pm$ 0.023 & 0 \\ 
			Centralized GAN & 0.943  & 3.031 $\pm$ 0.016 & 14.90 & 0.943  & 3.031 $\pm$ 0.016 & 14.90 \\ \midrule
			Avg GAN & 0.822  & 3.144 $\pm$ 0.013 & 41.63 & 0.936 & 2.877 $\pm$ 0.013  & 17.90 \\
			MD-GAN & 0.567 & 3.035 $\pm$ 0.011 & 56.19 & 0.936  & 2.951 $\pm$ 0.019  & 16.81   \\
			\midrule
			UA-GAN & 0.933 & 2.949 $\pm$ 0.023 & 20.80 & 0.923 & 2.875 $\pm$ 0.013 & 17.34\\
			\bottomrule
		\end{tabular}
% 			\vspace{-.05in}
	\end{center}
	\caption{The classification accuracy and IS, FID scores on mixture of three datasets. }	
	\label{tab:threedatasets}
% 	\vspace{-.2em}
\end{table}

\subsection{Empirical Results in larger federated learning setting}
\begin{figure}[ht]
    \centering
     \begin{minipage}{0.45\linewidth}
    \centering\includegraphics[width=\textwidth]{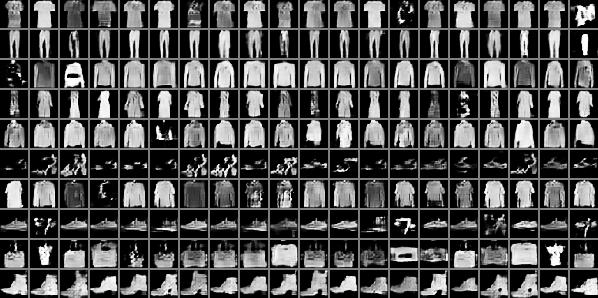} \\ (a) Avg-GAN MNIST+Fashion(50 data sites)
    \end{minipage}
    \begin{minipage}{0.45\linewidth}
    \centering\includegraphics[width=\textwidth]{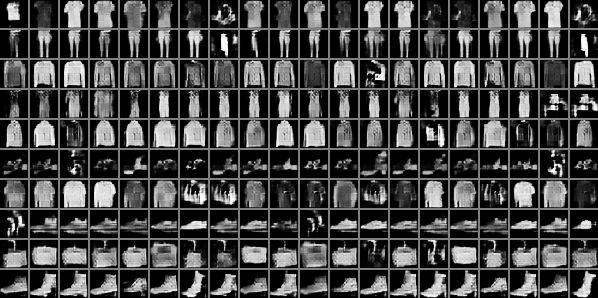}  \\ (b) MD-GAN MNIST+Fashion(50 data sites)
    \end{minipage}
    \begin{minipage}{0.45\linewidth}
    \centering\includegraphics[width=\textwidth]{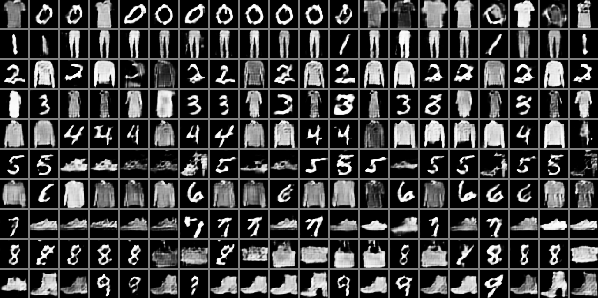} \\ (c)  UA-GAN MNIST+Fashion(50 data sites)
    \end{minipage}
    \caption{Synthetic images on the non-identical MNIST+Fashion dataset for 50 distributed sites using the average method, MD-GAN~\cite{hardy2019md} and our UA-GAN method.}
    \label{fig:50db}
\end{figure}

We report the results when using larger scale nodes($n=50$) in distributed GAN methods(Avg-GAN, MD-GAN and UA-GAN). We uniformly split each individual site of the non-identical MNIST + Fashion dataset into 5 distributed sites. In total, we adopt $50$ non-identical MNIST+Fashion datasets with 2380 MNIST and Fashion images each. The quantitative results are shown in Table~\ref{tab:50db}, and the synthetic images are shown in Fig~\ref{fig:50db}.

\begin{table}[!ht]
	\begin{center}
% 	\vspace{-.1em}
		\begin{tabular}{l|ccc|ccc}
		    \midrule
		    \multirow{2}{*}{Dataset} 
		    & \multicolumn{3}{c|}{non-i.i.d MNIST + Fashion (50 data sites)} \\ 
		    \cmidrule{2-4}
		    & Accuracy$\uparrow$ & IS$\uparrow$ & FID$\downarrow$ \\ 
		    \midrule
			Real & 0.943 & 3.620 $\pm$ 0.021 & 0  \\ 
			Centralized GAN & 0.904  & 3.437 $\pm$ 0.021 & 8.35  \\ \midrule
			Avg GAN & 0.489  & 3.755 $\pm$ 0.023 & 90.36  \\
			MD-GAN & 0.465 & 3.830 $\pm$ 0.020  & 89.36  \\
			\midrule
			UA-GAN & 0.626 & 3.531 $\pm$ 0.018  & 53.26 \\
			\bottomrule
		\end{tabular}
% 			\vspace{-.05in}
	\end{center}
	\caption{The classification accuracy and IS, FID scores on non-i.i.d mixture datasets for $50$ distributed sites. }	
	\label{tab:50db}
% 	\vspace{-.2em}
\end{table}

\subsection{Empirical Results in unconditional setting}
\begin{figure}[ht]
    \centering
     \begin{minipage}{0.45\linewidth}
    \centering\includegraphics[width=\textwidth]{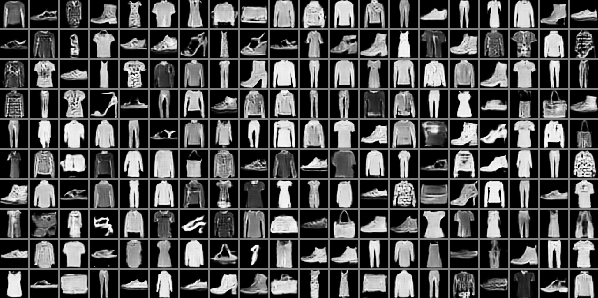} \\ (a) Avg GAN MNIST+Fashion
    \end{minipage}
    \begin{minipage}{0.45\linewidth}
    \centering\includegraphics[width=\textwidth]{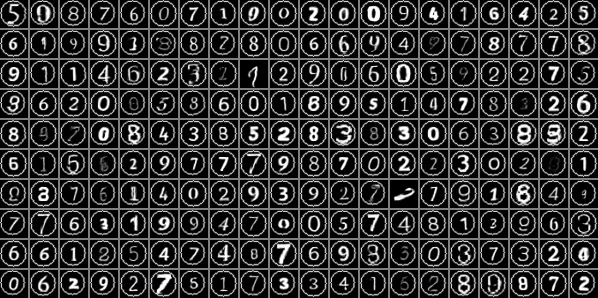}  \\ (b) Avg GAN MNIST+Font
    \end{minipage}
    \begin{minipage}{0.45\linewidth}
    \centering\includegraphics[width=\textwidth]{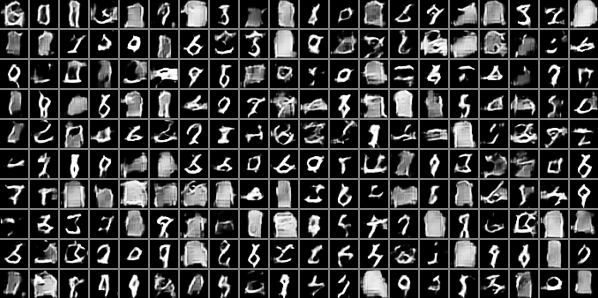} \\ (c)  MD-GAN MNIST+Fashion
    \end{minipage}
    \begin{minipage}{0.45\linewidth}
    \centering\includegraphics[width=\textwidth]{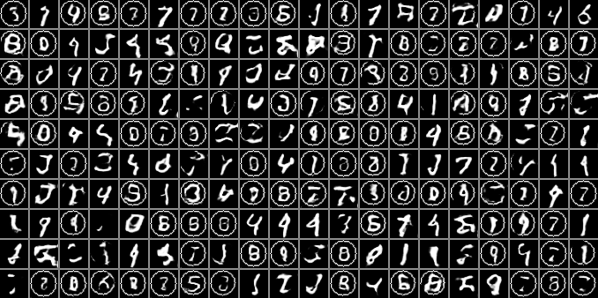}  \\ (d) MD-GAN MNIST+Font
    \end{minipage}
    \begin{minipage}{0.45\linewidth}
    \centering\includegraphics[width=\textwidth]{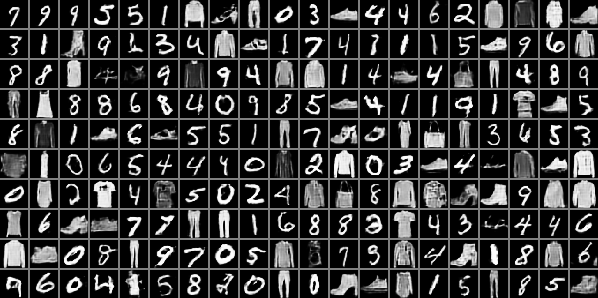}  \\ (e) UA-GAN MNIST+Fashion
    \end{minipage}
    \begin{minipage}{0.45\linewidth}
    \centering\includegraphics[width=\textwidth]{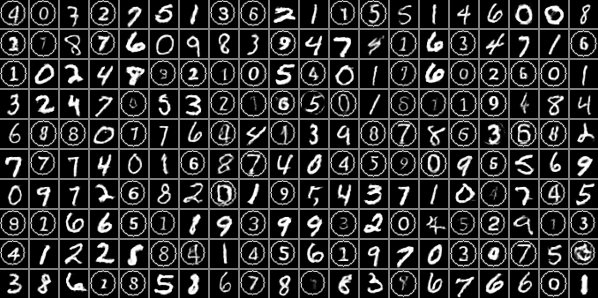} \\ (f) UA-GAN  MNIST+Font
    \end{minipage}
    \caption{Synthetic images on the non-identical MNIST+Fashion dataset ((a),(c),(e)) and MNIST+Font dataset ((b),(d),(f)) in unconditional setting.}
    \label{fig:unconditional-noniid}
    % \vspace{-1em}
\end{figure}

\begin{figure}[ht]
    \centering
     \begin{minipage}{0.45\linewidth}
    \centering\includegraphics[width=\textwidth]{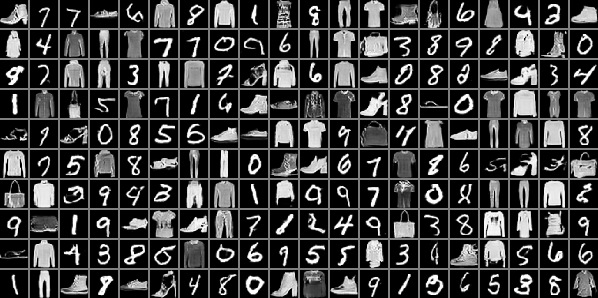} \\ (a) Avg GAN MNIST+Fashion
    \end{minipage}
    \begin{minipage}{0.45\linewidth}
    \centering\includegraphics[width=\textwidth]{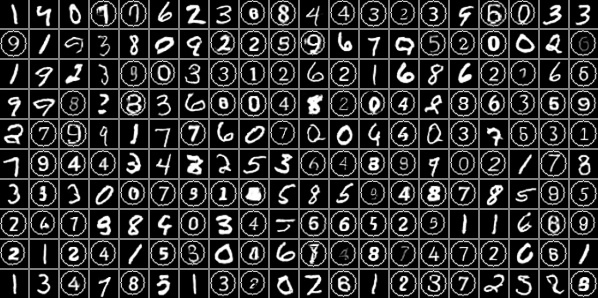}  \\ (b) Avg GAN MNIST+Font
    \end{minipage}
    \begin{minipage}{0.45\linewidth}
    \centering\includegraphics[width=\textwidth]{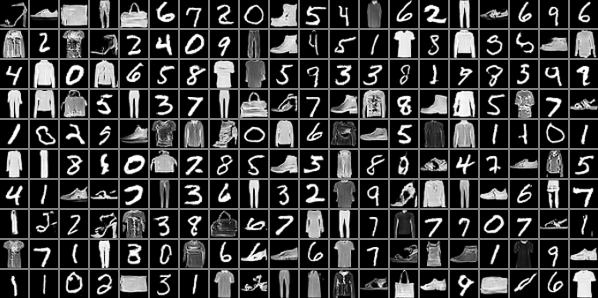} \\ (c)  MD-GAN MNIST+Fashion
    \end{minipage}
    \begin{minipage}{0.45\linewidth}
    \centering\includegraphics[width=\textwidth]{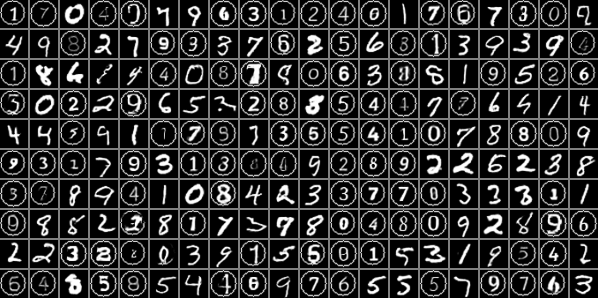}  \\ (d) MD-GAN MNIST+Font
    \end{minipage}
    \begin{minipage}{0.45\linewidth}
    \centering\includegraphics[width=\textwidth]{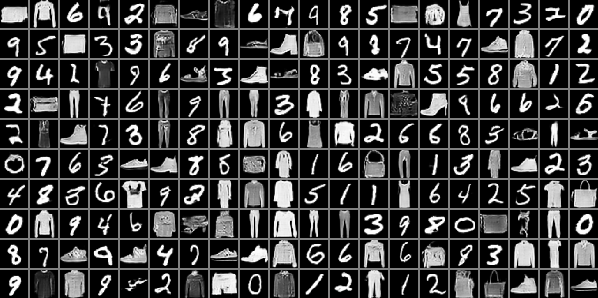}  \\ (e) UA-GAN MNIST+Fashion
    \end{minipage}
    \begin{minipage}{0.45\linewidth}
    \centering\includegraphics[width=\textwidth]{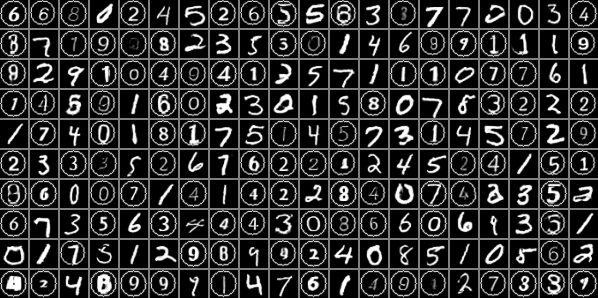} \\ (f) UA-GAN  MNIST+Font
    \end{minipage}
    \caption{Synthetic images on the identical MNIST+Fashion dataset ((a),(c),(e)) and MNIST+Font dataset ((b),(d),(f)) in unconditional setting.}
    \label{fig:unconditional-iid}
    % \vspace{-1em}
\end{figure}

We report the results when using unconditional GAN in all methods (centralized, Avg-GAN, MD-GAN and UA-GAN). The quantitative results are shown in Table~\ref{tab:unconditional-noniid} and Table~\ref{tab:unconditional-iid}. The synthetic images are shown in Fig.~\ref{fig:unconditional-noniid} and Fig.~\ref{fig:unconditional-iid}. In the unconditional setting, the condition variable (labels) won't be given thus one can not directly apply the synthetic data in training classification model.  Therefore we don't compute the classification accuracy in Table~\ref{tab:unconditional-noniid} and Table~\ref{tab:unconditional-iid}.

\begin{table}[!ht]
	\begin{center}
	\vspace{-.1em}
		\begin{tabular}{l|ccc|ccc}
		    \midrule
		    \multirow{2}{*}{Dataset} 
		    & \multicolumn{3}{c|}{non-i.i.d MNIST + Fashion}
		    & \multicolumn{3}{c}{non-i.i.d MNIST + Font} \\ 
		    \cmidrule{2-7}
		    & Accuracy$\uparrow$ & IS$\uparrow$ & FID$\downarrow$ & Accuracy$\uparrow$ & IS$\uparrow$ & FID$\downarrow$ \\ 
		    \midrule
			Real & 0.943 & 3.620 $\pm$ 0.021 & 0 & 0.994 & 2.323 $\pm$ 0.011 & 0 \\ 
			Centralized GAN & - & 3.387 $\pm$ 0.019 & 8.46 &  -  & 1.975 $\pm$ 0.009 & 17.56 \\ \midrule
			Avg GAN & - & 4.068 $\pm$ 0.020 & 61.56 & - & 1.547 $\pm$ 0.005  & 80.07 \\
			MD-GAN & - & 2.852 $\pm$ 0.021  & 60.34 & -  & 1.887 $\pm$ 0.007  & 36.36   \\
			\midrule
			UA-GAN & - & 3.280 $\pm$ 0.022 & 22.34 & - & 1.985 $\pm$ 0.013 & 22.17\\
			\bottomrule
		\end{tabular}
% 			\vspace{-.05in}
	\end{center}
	\caption{The classification accuracy and IS, FID scores on two non-i.i.d mixture datasets in unconditional setting. }	
	\label{tab:unconditional-noniid}
% 	\vspace{-.2em}
\end{table}

\begin{table}[!ht]
	\begin{center}
% 	\vspace{-.1em}
		\begin{tabular}{l|ccc|ccc}
		    \midrule
		    \multirow{2}{*}{Dataset} 
		    & \multicolumn{3}{c|}{i.i.d MNIST + Fashion}
		    & \multicolumn{3}{c}{i.i.d MNIST + Font} \\ 
		    \cmidrule{2-7}
		    & Accuracy$\uparrow$ & IS$\uparrow$ & FID$\downarrow$ & Accuracy$\uparrow$ & IS$\uparrow$ & FID$\downarrow$ \\ 
		    \midrule
			Real & 0.943 & 3.620 $\pm$ 0.021 & 0 & 0.994 & 2.323 $\pm$ 0.011 & 0 \\ 
			Centralized GAN & - & 3.387 $\pm$ 0.019 & 8.46 & -  & 1.975 $\pm$ 0.009 & 17.56 \\ \midrule
			Avg GAN & - & 3.326 $\pm$ 0.016 & 9.68 & - &  1.918 $\pm$ 0.006  & 19.52 \\
			MD-GAN & - & 3.428 $\pm$ 0.025  & 12.04 & -  & 1.934 $\pm$ 0.006   &  18.85  \\
			\midrule
			UA-GAN & -  & 3.367 $\pm$ 0.018 & 9.99 & - & 1.937 $\pm$ 0.009 & 18.83\\
			\bottomrule
		\end{tabular}
			\vspace{-.05in}
	\end{center}
	\caption{The classification accuracy and IS, FID scores on two i.i.d mixture datasets in unconditional setting. }	
	\label{tab:unconditional-iid}
% 	\vspace{-.2em}
\end{table}

\subsection{Empirical results of imbalanced datasets in different sites}
\begin{figure}[ht]
    \centering
     \begin{minipage}{0.45\linewidth}
    \centering\includegraphics[width=\textwidth]{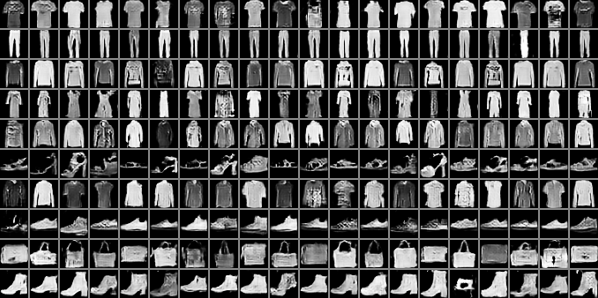} \\ (a) Avg-GAN MNIST+Fashion
    \end{minipage}
    \begin{minipage}{0.45\linewidth}
    \centering\includegraphics[width=\textwidth]{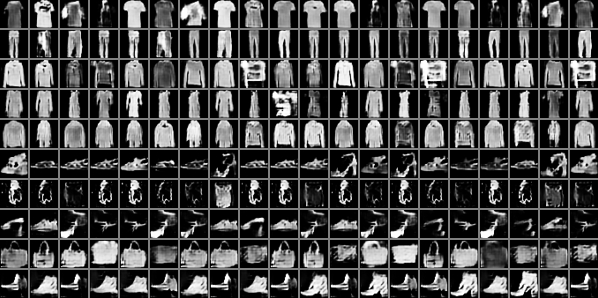}  \\ (b) MD-GAN MNIST+Fashion
    \end{minipage}
    \begin{minipage}{0.45\linewidth}
    \centering\includegraphics[width=\textwidth]{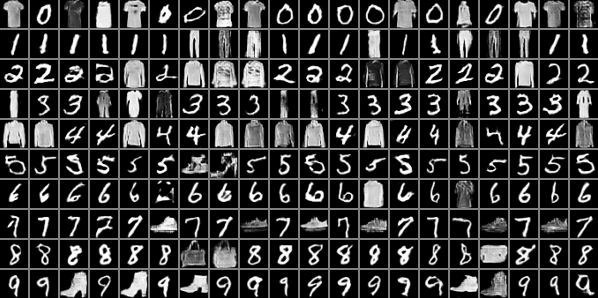} \\ (c)  UA-GAN MNIST+Fashion
    \end{minipage}
    \caption{Synthetic images on the imbalanced non-identical MNIST+Fashion dataset.}
    \label{fig:imbalanced-noniid}
    % \vspace{-1em}
\end{figure}

We report the results when the sizes of the 10 sites are not the same. Based on the non-identical MNIST + fashionMNIST dataset, we reduce the sample sizes of the first 5 sites by half and keep the other 5 sites unchanged. In this case, the numbers of images in each site are shown in Table~\ref{tab:imbalanced-data}. The quantitative results are shown in Table~\ref{tab:imbalanced-noniid}. The synthetic images are shown in Fig.~\ref{fig:imbalanced-noniid}.

\begin{table}[]
    \centering
    \begin{tabular}{c|c|c|c|c|c|c|c|c|c|c}
    \toprule
         & $D_0$ & $D_1$ & $D_2$ & $D_3$ & $D_4$ & $D_5$ & $D_6$ & $D_7$ & $D_8$ & $D_9$ \\ \midrule
         MNIST & 2917 & 3393 & 2944 & 3032 & 2939 & 5421 & 5918 & 6265 & 5851 & 5949 \\
         Fashion & 3044 & 2978 & 3035 & 3033 & 2982 & 6000 & 6000 & 6000 & 6000 & 6000 \\
         Total & 5961 & 6371 & 5979 & 6065 & 5921 & 11421 & 11918 & 12265 & 11851 & 11949 \\
         \bottomrule
    \end{tabular}
    \caption{The image numbers in each site in the imbalanced setting.}
    \label{tab:imbalanced-data}
\end{table}

\begin{table}[!ht]
	\begin{center}
% 	\vspace{-.1em}
		\begin{tabular}{l|ccc}
		    \midrule
		    & Accuracy$\uparrow$ & IS$\uparrow$ & FID$\downarrow$ \\ 
		    \midrule
			Real & 0.939 & 3.580 $\pm$ 0.039 & 0  \\ 
			Centralized GAN & 0.886 & 3.486 $\pm$ 0.033 & 10.87 \\ \midrule
			Avg GAN & 0.497 & 3.809  $\pm$ 0.025 & 74.45  \\
			MD-GAN & 0.443 & 3.877 $\pm$ 0.034 & 85.61  \\
			\midrule
			UA-GAN & 0.846 & 2.717 $\pm$ 0.019 & 30.30 \\
			\bottomrule
		\end{tabular}
	\end{center}
	\caption{The classification accuracy and IS, FID scores on the imbalanced non-i.i.d mixture MNIST + fashionMNIST dataset. }	
	\label{tab:imbalanced-noniid}
% 	\vspace{-.2em}
\end{table}

\end{document}